\documentclass[11pt,a4paper]{article}
\usepackage{booktabs}
\usepackage[utf8]{inputenc}
\usepackage{geometry}
\usepackage{graphicx}
\usepackage{multirow}
\usepackage{amsmath}
\usepackage{amsthm}
\usepackage{stackengine}
\usepackage{amssymb}
\usepackage{algorithm}
\usepackage{algorithmicx}
\usepackage{algpseudocode} 
\usepackage{makecell}
\usepackage{color}
\usepackage{comment}

\usepackage{hyperref}
\hypersetup{
    colorlinks=true,
    linkcolor=blue,
    filecolor=blue,      
    urlcolor=blue,
    citecolor=blue
}

\newtheorem{mytheorem}{Theorem}
\newtheorem{mylemma}{Lemma}
\newtheorem{definition}{Definition}

\newcommand{\Input}{{\hspace*{\algorithmicindent} \textbf{Input }}}

\newcommand{\Output}{{\hspace*{\algorithmicindent} \textbf{Output }}}

\algnewcommand{\LineComment}[1]{\State \(\triangleright\) #1}

\title{The Tsetlin Machine -- A Game Theoretic Bandit Driven Approach to Optimal Pattern Recognition with\\Propositional Logic\footnote{Source code and datasets for this paper can be found at \href{https://github.com/cair/TsetlinMachine}{https://github.com/cair/TsetlinMachine} and \href{https://github.com/cair/pyTsetlinMachine}{https://github.com/cair/pyTsetlinMachine}.}}
\author{Ole-Christoffer Granmo\thanks{Author's status: {\it Professor}. The author can be contacted at: Centre for Artificial Intelligence Research (\href{https://cair.uia.no}{https://cair.uia.no}), University of Agder, Grimstad, Norway.  E-mail: {\tt ole.granmo@uia.no}}}
\date{}

\begin{document}

\maketitle

\begin{abstract}
Although simple individually, artificial neurons provide state-of-the-art performance when interconnected in deep networks. Arguably, the Tsetlin Automaton is an even simpler and more versatile learning mechanism, capable of solving the multi-armed bandit problem. Merely by means of a single integer as memory, it learns the optimal action in stochastic environments through increment and decrement operations. In this paper, we introduce the Tsetlin Machine, which solves complex pattern recognition problems with propositional formulas, composed by a collective of Tsetlin Automata. To eliminate the longstanding problem of vanishing signal-to-noise ratio, the Tsetlin Machine orchestrates the automata using a novel game. Further, both inputs, patterns, and outputs are expressed as bits, while recognition and learning rely on bit manipulation, simplifying computation. Our theoretical analysis establishes that the Nash equilibria of the game align with the propositional formulas that provide optimal pattern recognition accuracy. This translates to learning without local optima, only global ones. In five benchmarks, the Tsetlin Machine provides competitive accuracy compared with SVMs, Decision Trees, Random Forests, Naive Bayes Classifier, Logistic Regression, and Neural Networks. We further demonstrate how the propositional formulas facilitate interpretation. In conclusion, we believe the combination of high accuracy, interpretability, and computational simplicity makes the Tsetlin Machine a promising tool for a wide range of domains. 
\end{abstract}

\section{Introduction}
 
Although simple individually, artificial neurons provide state-of-the-art performance when interconnected in deep networks \cite{LeCun2015}.
However, deep neural networks often require huge amounts of training data and extensive computational resources. Unknown to many, there exists an arguably even more fundamental and versatile learning mechanism than the artificial neuron, namely, the Tsetlin Automaton, developed by M.L. Tsetlin in the Soviet Union in the early 1960s \cite{Tsetlin1961}. In this paper, we propose a novel technique for large-scale and complex pattern recognition based on Tsetlin Automata.

\subsection{The Tsetlin Automaton}\label{sec:tsetlin_automaton}
Tsetlin Automata have been used to model biological systems, and have attracted considerable interest because they can learn the optimal action when operating in unknown stochastic
environments \cite{Tsetlin1961,Narendra1989}. Furthermore, they combine rapid and accurate convergence with low computational complexity.

In all brevity, the Tsetlin Automaton is one of the pioneering solutions to the well-known multi-armed bandit problem \cite{Robbins1952,Gittins1979} and the first Learning Automaton \cite{Narendra1989}. It performs actions $\alpha_z$, $z\in \{1, 2\}$, sequentially in an environment. Each action $\alpha_z$ triggers either a reward or a penalty. That is, the action is rewarded with probability $p_z$, otherwise it is penalized. The reward  probabilities are unknown to the automaton and may change over time. Under such challenging conditions, the goal is to identify the action with the highest reward probability using as few attempts as possible.

\begin{figure}[!ht]
\centering
\includegraphics[width=5.0in]{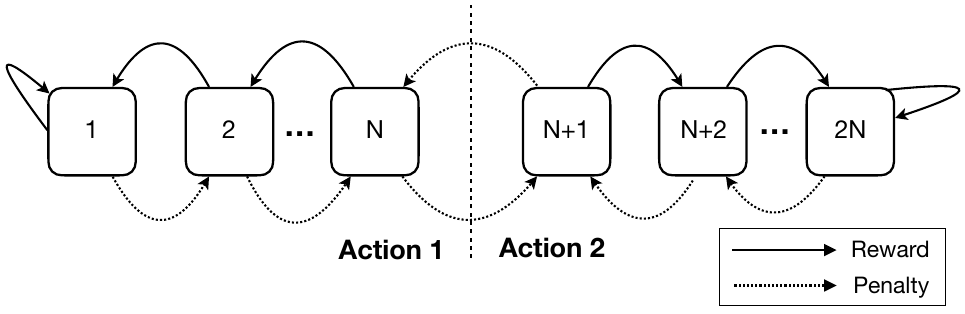}
\caption{A Tsetlin Automaton for two-action environments.}
\label{figure:tsetlin_automaton}
\end{figure}
The mechanism driving the Tsetlin Automaton is surprisingly simple. Informally, as illustrated in Figure \ref{figure:tsetlin_automaton}, a Tsetlin Automaton is simply a fixed finite-state automaton \cite{Carroll1989} with an unusual interpretation:
\begin{itemize}
\item The current state of the automaton decides which action to perform. The automaton in the figure has $2N$ states. Action 1 ($\alpha_1$) is performed in the states with index $1$ to $N$, while Action 2 ($\alpha_2$) is performed in the states with index $N+1$ to $2N$.
\item The state transitions of the automaton govern learning. One set of state transitions is activated on reward (solid lines in the figure), and one set of state transitions is activated on penalty (dotted lines in the figure). As seen, rewards and penalties trigger specific transitions from one state to another, designed to reinforce successful actions (those eliciting rewards). 
\end{itemize}
Formally, a Two-Action Tsetlin Automaton can be defined as a quintuple \cite{Narendra1989}:
\[
\{\underline{\Phi}, \underline{\alpha}, \underline{\beta}, F(\cdot,\cdot), G(\cdot)\}.
\]
$\underline{\Phi} = \{\phi_1, \phi_2, …, \phi_{2N}\}$ is the set of internal states. $\underline{\alpha} = \{\alpha_1, \alpha_2\}$ is the set of automaton actions. $\mathbf{\underline{\beta}} = \{\beta_{\mathrm{Penalty}}, \beta_{\mathrm{Reward}}\}$ is the set of inputs that can be given to the automaton.
An output function $G(\phi_u)$ determines the next action performed by the automaton, 
given the current automaton state $\phi_u$: 
\begin{equation}
G(\phi_u) = \begin{cases}
       \alpha_1, & \mathbf{if}~ 1 \le u \le N\\
       \alpha_2, & \mathbf{if}~ N+1 \le u \le 2N.
     \end{cases}\label{eqn:action_mapping}
\end{equation}

Finally, a transition function $F(\phi_u, \beta_v)$ determines the new automaton state from: (i) the current automaton state $\phi_u$  and (ii) the response $\beta_v$ of the environment to the action performed by the automaton: 
\begin{equation}
F(\phi_u, \beta_v) = \begin{cases}
     \phi_{u+1},& \mathbf{if}~ 1 \le u \le N ~\mathbf{and}~ v = \text{Penalty}\\
     \phi_{u-1},& \mathbf{if}~ N+1 \le u \le 2N ~\mathbf{and}~ v = \text{Penalty}\\
     \phi_{u-1},& \mathbf{if}~ 1 < u \le N ~\mathbf{and}~ v = \text{Reward}\\
     \phi_{u+1},& \mathbf{if}~ N+1 \le u < 2N ~\mathbf{and}~ v = \text{Reward}\\
     \phi_{u},& \mathbf{otherwise}.
     \end{cases}\label{eqn:state_transition}
\end{equation}

Implementation-wise, a Tsetlin Automaton simply maintains an integer (the state index), and learning is performed through increment and decrement operations, according to the transitions specified by $F(\phi_u, \beta_v)$ (and depicted in Figure \ref{figure:tsetlin_automaton}). \emph{The Tsetlin Automaton is thus extremely simple computationally, with a very small memory footprint.}

\subsection{State-of-the-art in the Field of Learning Automata}

Learning Automata have attracted considerable interest because they can learn the optimal action when operating in unknown stochastic environments \cite{Narendra1989}. As such, they have been used for pattern classification over more than four decades. Early work includes the stochastic Learning Automata-based classifier of Barto and Anandan \cite{barto1985pattern}, as well as the games of Learning Automata proposed by Narendra and Thathachar \cite{Narendra1989}. These approaches can learn the optimal classifier for specific forms of discriminant functions (e.g., linear classifiers), also when feedback is noisy. Along the same lines, Sastry and Thathachar have provided several algorithms based on cooperating systems of Learning Automata \cite{sastry1999learning}. 
More recently, Zahiri proposed hyperplane- and rule-based classifiers, which performed comparably to other well-known methods in the literature \cite{zahiri2008learning,zahiri2012classification}. Recent research also includes the noise-tolerant learning algorithm by Sastry et al., built upon a team of continuous-action Learning Automata \cite{sastry2009team}. Further, Goodwin et al. proposed a Learning Automata-guided random walk on a grid to construct robust discriminant functions \cite{goodwin2016distributed}, while Motieghader et al. introduced a hybrid scheme that combines a genetic algorithm with Learning Automata for classification \cite{motieghader2017hybrid}.

In general, however, previous Learning Automata schemes have mainly addressed relatively small-scale pattern recognition tasks. The above solutions are further primarily based on so-called \emph{variable structure learning automata} \cite{Thathachar2004}. Although still simple, these are significantly more complex than the Tsetlin Automaton because they need to maintain an action probability vector for sampling actions. Their success in pattern recognition is additionally limited by constrained pattern representation capability (linearly separable classes and simple decision trees). 

The Tsetlin Automaton has formed the core of more advanced learning automata designs as well. This includes decentralized control \cite{Tung1996}, searching on the line \cite{Oommen1997}, equi-partitioning \cite{Oommen1988}, streaming sampling for social activity networks \cite{Ghavipour2018}, faulty dichotomous search \cite{Yazidi2018}, learning in deceptive environments \cite{Zhang2016a}, and routing in telecommunication networks \cite{Oommen2007c}. The strength of these finite state learning automata solutions is that they have provided state-of-the-art performance when problem properties are unknown and stochastic, while the problem must be solved as quickly as possible through trial and error.

\subsection{The Vanishing Signal-to-Noise Ratio Problem}\label{sec:vanishing}

The ability to handle stochastic and unknown environments for a wide range of problems, combined with their computational simplicity and small memory footprint, make Learning Automata an attractive building block for complex machine learning tasks. However, a particularly adverse challenge has hindered the success of Learning Automata. Complex problem solving requires a team of interacting automata, and unfortunately, each team member introduces noise. This challenge manifests sharply in the Goore Game, which employs a simple uni-modal stochastic objective function to be optimized by a Learning Automata team~\cite{Granmo2013a,Tung1996}.

The challenge manifests as follows. Consider a team of $W+1$ two-action variable structure Learning Automata. Each automaton maintains an action probability $p$. In any round of the Goore Game, an automaton performs its first action with probability $p$ and its second action with probability $1-p$. Now, let $\mu$ be the average value with which a single automaton can modify the game's objective function by switching action. From the perspective of a single automaton, each other automaton in the team then increases the objective function variance, amounting to $p(1-p) \mu
^2$ \cite{Granmo2013a}. In particular, consider the case of a deterministic objective function, with $p$ being the action probability within the team that currently is farthest away from $0.5$. Then the best-case signal-to-noise ratio (SNR) becomes:
\begin{eqnarray}
\mathrm{SNR} = \frac{\mu^2}{\sigma^2} = \frac{\mu^2}{W p(1-p)\mu^2} = \frac{1}{W p(1-p)}.
\end{eqnarray}
That is, as the number of automata $W$ grows, SNR drops to zero (vanishes).

For Tsetlin Automata, which are deterministic, the stochasticity of the action selection arises from the automata changing state. These state changes are stochastic since they are driven by stochastic feedback. Thus, again, as the number of automata grows, so does the variance of the feedback. Even for a simple uni-modal objective function, such as the one used in the Goore Game, this effect is problematic. Indeed, as explored by Kleinrock and Tung in 1996, a team of Tsetlin Automata solving the Goore Game needs to maintain an increasingly larger state-space as the number of Tsetlin Automata grows, and in the end an infinite number of states per Tsetlin Automaton \cite{Tung1996}.

In general, this problem is inherent to decision-making with Learning Automata because of their decentralized and stochastic nature. The automata independently decide upon their actions, directly based on the feedback from the environment. This is on one hand a strength because it allows problems to be solved in a decentralized manner. On the other hand, as the number of automata grows, the level of noise increases. In the following, we will refer to this effect as the \emph{vanishing signal-to-noise ratio problem}.

\begin{table}[!bh]
    \centering
    \begin{tabular}{cccccccc}
            0 0 * 1 * 0 0 0\\
            * 0 * 1 * 0 0 0\\
            0 * * 1 * * * 0\\
            0 * * * * 0 0 *\\
            0 0 0 * * 0 0 0\\
            0 * 0 * * * 0 0\\
            0 0 * 1 * * * 0\\
            0 0 0 * 1 * * *\\
    \end{tabular}
    \caption{A bit pattern produced by the Tsetlin Machine for handwritten digits '1'. The '*' symbol can either take the value '0' or '1'. The remaining bit values require strict matching. The pattern is relatively easy to interpret for humans compared to, e.g., the weights of a neural network. It is also efficient to evaluate for computers. Despite this simplicity, the Tsetlin Machine produces bit patterns that deliver competitive pattern recognition accuracy for several datasets, as reported in Section \ref{sec:empirical_results}.}
    \label{tab:bit_pattern}
\end{table}

\subsection{Paper Contributions}
In this paper, we attack the limited pattern expression capability and vanishing signal-to-noise ratio of Learning Automata-based pattern recognition, introducing the \emph{Tsetlin Machine}. The contributions of the paper can be summarized as follows:
\begin{itemize}
\item We introduce the Tsetlin Machine, which  solves complex pattern recognition problems with \emph{propositional formulas}, composed by a collective of Tsetlin Automata.
\item We eliminate the longstanding vanishing signal-to-noise ratio problem with a unique decentralized learning scheme based on game theory \cite{VonNeumann1947,Tsetlin1961}. The game we have designed allows \emph{millions} of Tsetlin Automata to successfully cooperate.
\item The game orchestrated by the Tsetlin Machine is based on resource allocation principles \cite{Granmo2007a}, in inter-play with frequent pattern mining \cite{Haugland2014}. By allocating sparse pattern representation resources according to the frequency of the patterns, the Tsetlin Machine is able to capture intricate unlabelled sub-patterns, for instance addressing the so-called Noisy XOR-problem.
\item Our theoretical analysis establishes that the Nash equilibria of the Tsetlin Machine game are aligned with the propositional formulas that provide optimal pattern recognition accuracy. This translates to learning without local optima, only global ones.
\item The propositional formulas are represented as bit patterns. These bit patterns are relatively easy to interpret, compared to e.g. a neural network (see Table \ref{tab:bit_pattern} for an example bit pattern). This facilitates human quality assurance and scrutiny, which for instance can be important in safety-critical domains such as medicine.
\item The Tsetlin Machine is a new approach to global construction of decision rules \cite{GuolongSu2016,Wang2017}. We demonstrate that decision rules for large-scale pattern recognition can be learned on-line, under particularly noisy conditions.
\item The Tsetlin Machine is particularly suited for digital computers, being directly based on bit manipulation with AND-, OR-, and NOT operators.
\item In our empirical evaluation on five datasets, the Tsetlin Machine provides competitive performance in comparison with Multilayer Perceptron Networks, Support Vector Machines, Decision Trees, Random Forests, the Naive Bayes Classifier, and Logistic Regression.
\item We demonstrate how the Tsetlin Machine can be used as a building block to create more advanced architectures.
\end{itemize}
In conclusion, we believe the combination of high accuracy, interpretability, and computational simplicity makes the Tsetlin Machine a promising tool for a wide range of domains.

\subsection{Paper Organization}

The paper is organized as follows. In Section \ref{sec:pattern_recognition_problem}, we define the exact nature of the pattern recognition problem we are going to solve, also introducing the crucial concept of sub-patterns.

Then, in Section \ref{sec:tsetlin_machine}, we cover the Tsetlin Machine in detail. We first present the Tsestlin Machine inference structure, before we introduce the Tsetlin Automata teams that write formulas in propositional logic. These Tsetlin Automata teams are in turn organized to recognize complex patterns. We conclude the section by presenting the Tsetlin Machine game that we use to coordinate millions of Tsetlin Automata, eliminating the vanishing signal-to-noise ratio problem.

In Section \ref{sec:theoretical_results}, we analyze pertinent properties of the Tsetlin Machine formally, and establish that the Nash equilibria of the game are aligned with the propositional formulas that solve the pattern recognition problem at hand. This allows the Tsetlin Machine as a whole to robustly and accurately uncover complex patterns with propositional logic.

In our empirical evaluation in Section \ref{sec:empirical_results}, we evaluate the performance of the Tsetlin Machine on five datasets: Flower categorization, digit recognition, board game planning, the Noisy XOR Problem with Non-informative Features, as well as MNIST.

The Tsetlin Machine has been designed to act as a building block in more advanced architectures, and in Section \ref{sec:building_block} we demonstrate how four distinct architectures can be built by interconnecting multiple Tsetlin Machines.

As the first step in a new research direction, the Tsetlin Machine also opens up a range of new research questions. In Section \ref{sec:conclusion}, we summarize our main findings and provide pointers to some of the open research problems.

\section{The Pattern Recognition Problem}
\label{sec:pattern_recognition_problem}

\begin{figure}[!t]
\centering
\includegraphics[width=3.0in]{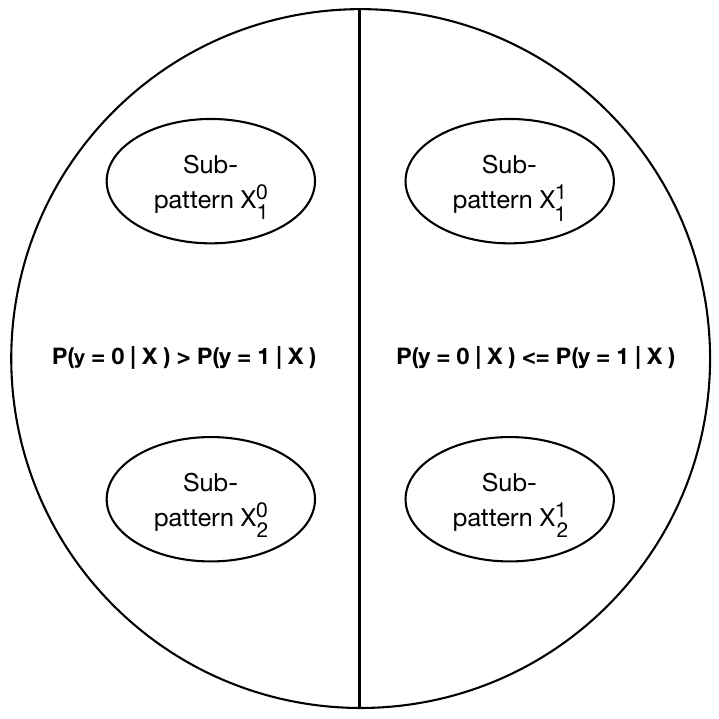}
\caption{A partitioning of the input space according to the posterior probability of the output variable $y$, highlighting distinct sub-patterns in the input space, $X \in \mathcal{X}$. Sub-patterns most likely belonging to output $y = 1$ can be found on the right side, while sub-patterns most likely belonging to $y = 0$ on the left.}
\label{figure:classification_problem}
\end{figure}

The accuracy of a machine learning technique is bounded by its pattern representation capability. The Naive Bayes Classifier, for instance, assumes that input variables are independent given the output category. In general, compared to the representation capability of the underlying language of digital computers, namely, Boolean algebra, most machine learning techniques appear somewhat limited, neural networks being one of the exceptions. With propositional logic/Boolean algebra\footnote{We found the Tsetlin Machine on propositional logic, which can be mapped to Boolean algebra, and vice versa.} as the starting point, we here define the pattern recognition problem to be solved by the Tsetlin Machine.

\paragraph{Input and Output.} Consider an input vector of $o$ propositional variables: $X=(x_1,\ldots,x_o) \in \mathcal{X}, \mathcal{X} = \{0,1\}^{o}$. Together with their negated counterparts, $\bar{x}_k  = \lnot x_k = 1-x_k$, the variables form a literal set $L = \{l_1, \ldots, l_{2o}\} = \{x_1,\ldots,x_o, \bar{x}_1,\ldots, \bar{x}_o\}$. From the literals, we are to produce an output vector of $m$ propositional variables: $Y = (y^1, y^2, \ldots, y^m) \in \mathcal{Y}, \mathcal{Y} = \{0,1\}^{m}$. For the sake of simple notation, we will in the remainder of the paper represent any particular output variable $y^i, i=1,\ldots,m$, as a single output variable $y$, without loss of generality.

\paragraph{Patterns.} In Tsetlin Machine learning, a pattern is expressed as a conjunctive clause in propositional logic, denoted $C_j$, with $j$ being the index of the clause. A conjunctive clause is formed by ANDing a subset $L_j \subseteq L$ of the literal set:
\begin{eqnarray}
C_j(X)&=&\bigwedge_{l_k \in L_j} l_k.\label{eqn:clause}
\end{eqnarray}
or, equivalently, as a product of the literals:
\begin{eqnarray}
C_j(X)&=& \prod_{l_k \in L_j} l_k.
\end{eqnarray}
For example, the clause $C_j(X) = x_1 \land x_2 = x_1 x_2$ consists of the literals $L_j = \{x_1, x_2\}$ and outputs $1$ iff $x_1 = x_2 = 1$. 

Note that any formula in propositional logic can be reformulated as a disjunction of conjunctive clauses (referred to as disjunctive normal form). With $o$ input variables, one can express no less than $2^{2^o}$ unique propositional formulas, making conjunctive clauses a powerful building block for expressing complex non-linear patterns. Arguably, they are also particularly easy for humans to comprehend \cite{valiant1984learnable}.

\paragraph{Learning Problem and Sub-Patterns.}

As illustrated in Figure \ref{figure:classification_problem}, the input space $\mathcal{X}$ can be partitioned into two parts, $\mathcal{X}^1 = \{X | P(y = 0 | X) \le P(y = 1 | X)\}$ and $\mathcal{X}^0 = \{X | P(y = 0 | X) > P(y = 1 | X)\}$. For input vectors in $\mathcal{X}^1$, classification accuracy is maximized by assigning $y$ the value $1$. Conversely, for input vectors in $\mathcal{X}^0$,  accuracy is maximized by assigning $y$ the value $0$. \emph{Accordingly, the learning problem that we address consists of finding the partitioning of the input space that maximizes classification accuracy}.

We now come to the crucial concept of unlabelled sub-patterns. As Figure \ref{figure:classification_problem} exemplifies, we decompose $\mathcal{X}^0$ and $\mathcal{X}^1$ into $n/2$ sub-partitions each, respectively, $\mathcal{X}^0_j, j \in \{1, \ldots, n/2\},$ and $\mathcal{X}_j^1, j \in \{1, \ldots, n/2\}$. Each sub-partition $\mathcal{X}_j^\omega, \omega \in \{0,1\},$ is defined by a corresponding conjunctive clause, containing all the inputs that make the clause evaluate to $1$, and only those inputs. Consider for instance the XOR-relation. Then $\mathcal{X}^0$ decomposes into $\mathcal{X}^0_1 = \{X| x_1 x_2 = 1\}$ and $\mathcal{X}^0_2 = \{X| \bar{x}_1 \bar{x}_2 = 1\}$, while $\mathcal{X}^1$ decomposes into $\mathcal{X}_1^1 = \{X| x_1 \bar{x}_2 = 1\}$ and $\mathcal{X}_2^1 = \{X| \bar{x}_1 x_2 = 1\}$.

During pattern learning, however, we only observe training examples $(X, y)$ sampled from an unknown input-output distribution $P(X, y)$. Which sub-partition $X$ was sampled from is unavailable to us. What we know is that (i) each sub-partition can be described by a conjunctive clause, and (ii) the probability of obtaining a training example from a specific sub-partition is maximally close to a known value $\frac{1}{s}, s \ge 1$, but not equal to or below. That is, we are targeting sub-patterns with a certain frequency. Formally, the pattern recognition problem that we address can be defined as follows:
\begin{definition}\label{def:problem_definition}
Let $y$ be a random output variable taking values $\omega \in \{0,1\}$ and let $X=(x_1,\ldots,x_o)$ refer to $o$ random input variables with domain $\mathcal{X} = \{0,1\}^{o}$. Given the joint input-output distribution $P(X,y)$,  the input domain $\mathcal{X}$ partitions into two parts, $\mathcal{X}^1 = \{X | P(y = 0 | X) \le P(y = 1 | X)\}$ and $\mathcal{X}^0 = \{X | P(y = 0 | X) > P(y = 1 | X)\}$. Assume that the joint probability distribution $P(X, y)$ relates $y$  to $n$ sub-patterns in $\mathcal{X}$, $n/2$ sub-patterns per $y$-value.  Let each sub-pattern be defined by a corresponding conjunctive clause $Q_j^{\omega}(X)$, $j \in \{1,\ldots, n/2\}$, ${\omega} \in \{0,1\}$. That is, clause $Q_j^{\omega}(X)$ delineates a subset $\mathcal{X}_j^\omega$ of inputs within $\mathcal{X}$: $\mathcal{X}_j^\omega = \{X | Q_j^{\omega}(X) = 1\}$. Let these subsets have the following properties:
\begin{enumerate}
    \item Conditioned on $\mathcal{X}^\omega$, the probability of input subset $\mathcal{X}_j^\omega$ is larger than a given constant $\frac{1}{s}, s \in [1,\infty)$: $P(\mathcal{X}_j^\omega | \mathcal{X}^\omega) > \frac{1}{s}$.
    \item If clause $Q_j^{\omega}(X)$ is extended with additional literals, the resulting input subset $\mathcal{X}^{{\omega}'}_j$ becomes a subset of $\mathcal{X}^{\omega}_j$, $\mathcal{X}^{{\omega}'}_j \subset \mathcal{X}^{\omega}_j$, and occurs with probability less than $\frac{1}{s}$, conditioned on $\mathcal{X}^\omega$: $P(\mathcal{X}_j^{{\omega}'} | \mathcal{X}^\omega) < \frac{1}{s}$.
    \item Input outside the $\mathcal{X}^{\omega}_j$-subsets cannot satisfy property (2) above: $P(\mathcal{X} \setminus \bigcup_{j} \mathcal{X}^{\omega}_j | \mathcal{X}^\omega) < \frac{1}{s}$.
    \item Each input $X$ from the subset $\mathcal{X}^{\omega}_j$ is a noisy predictor of the associated $y$-value $\omega$, $P(y = \omega | \mathcal{X}_j^\omega) > 0.5$, implying $\mathcal{X}_j^\omega \subseteq \mathcal{X}^\omega$.
\end{enumerate}
Under these assumptions, given $n$ and $s$, the pattern recognition problem consists of recovering the clauses $Q_j^{\omega}(X)$, $j \in \{1,\ldots n/2\}, \omega \in \{0,1\}$, only by observing input-output samples $(X,y)$ from the unknown distribution $P(X,y)$. 
\end{definition}
Note that in practice, for real-world data, $n$ and $s$ may not be known. They can then be considered as hyper-parameters, optimized through a hyper-parameter search.

\section{The Tsetlin Machine}
\label{sec:tsetlin_machine}

The Tsetlin Machine decomposes problems into self-contained patterns that are expressed as conjunctive clauses in propositional logic. In this section, we first cover the clause-based  inference structure, before we proceed to present the Tsetlin Automata teams, which are responsible for composing the clauses. The teams are coordinated by means of a game that optimizes accuracy using so-called Type I and Type II Feedback, combined with resource allocation dynamics. We conclude the section with discussing the details of the game, including pseudo code.

\subsection{Tsetlin Machine Inference Structure}

\begin{figure}[!t]
\centering
\includegraphics[width=5.0in]{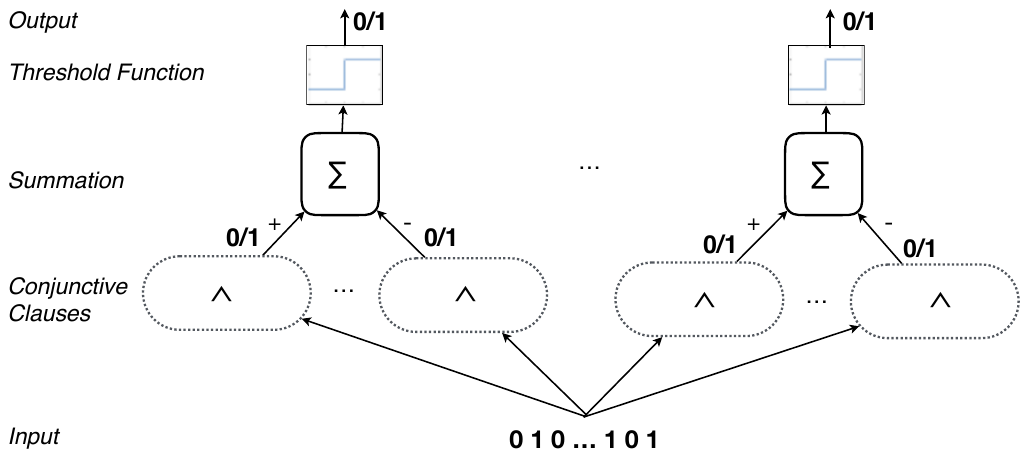}
\caption{The Tsetlin Machine inference structure, introducing clause polarity, a summation operator collecting "votes", and a threshold function arbitrating the final output.}
\label{figure:architecture_summation}
\end{figure}

The Tsetlin Machine inference structure is shown in Figure \ref{figure:architecture_summation}. As seen, an input vector $X=(x_1,\ldots,x_o) \in \mathcal{X}, \mathcal{X} = \{0,1\}^{o},$ of $o$ propositional variables is fed to several conjunctive clauses for evaluation.

The number of clauses employed is a user set parameter $n$. Half of the clauses are assigned positive polarity, denoted by upper index $1$: $C^1_j, j \in \{1, \ldots, n/2\}$. The other half is assigned negative polarity, denoted by upper index $0$: $C^0_j, j \in \{1, \ldots, n/2\}$. In the figure, polarity is indicated with a '+' or '-'  sign, attached to each clause.

The role of the clauses is to capture sub-patterns in $\mathcal{X}$ per Eq. \ref{eqn:clause}, however, with a slight adjustment. We allow clauses without literals, i.e., $L^\omega_j = \emptyset, j \in \{1, \ldots, n/2\}, \omega \in \{0, 1\}$. These output $1$ during learning and $0$ during classification:
\begin{eqnarray}
C^\omega_j(X) = \begin{cases}
1, & \textbf{if}~ L^\omega_j = \emptyset \text{~during learning},\\
0, & \textbf{if}~ L^\omega_j = \emptyset \text{~during classification},\\
\bigwedge_{l_k \in L^\omega_j} l_k,& \textbf{if}~ L^\omega_j \ne \emptyset.
\end{cases}
\end{eqnarray}
In other words, clauses without literals play no role during classification, and can be pruned from the inference structure. During learning, however, outputting $1$ stimulates further updating of the clause, as explored in Section \ref{sec:tsetlin_machine_game}.

The clause evaluations, in turn, are combined into a final output decision through summation and unit step thresholding, $u(v) = 1 ~\mathbf{if}~ v \ge 0 ~\mathbf{else}~ 0$:
\begin{equation}
\hat{y} = u\left(\sum_{j=1}^{n/2} C_j^1(X) -\sum_{j=1}^{n/2} C_j^0(X)\right).\label{eqn:output_function}
\end{equation}
That is, output is decided based on a majority vote, with the positive polarity clauses voting for $y=1$ and the negative ones for $y=0$. The purpose is to reach a balanced output decision, weighting positive against negative evidence, found in the input. The classifier
$\hat{y} = u(x_1 \bar{x}_2 + \bar{x}_1 x_2 - x_1 x_2 -$ $\bar{x}_1 \bar{x}_2)$, for instance, captures the XOR-relation.

\subsection{The Tsetlin Automata Team for Composing Clauses}

\begin{figure}[!t]
\centering
\includegraphics[width=0.85\textwidth]{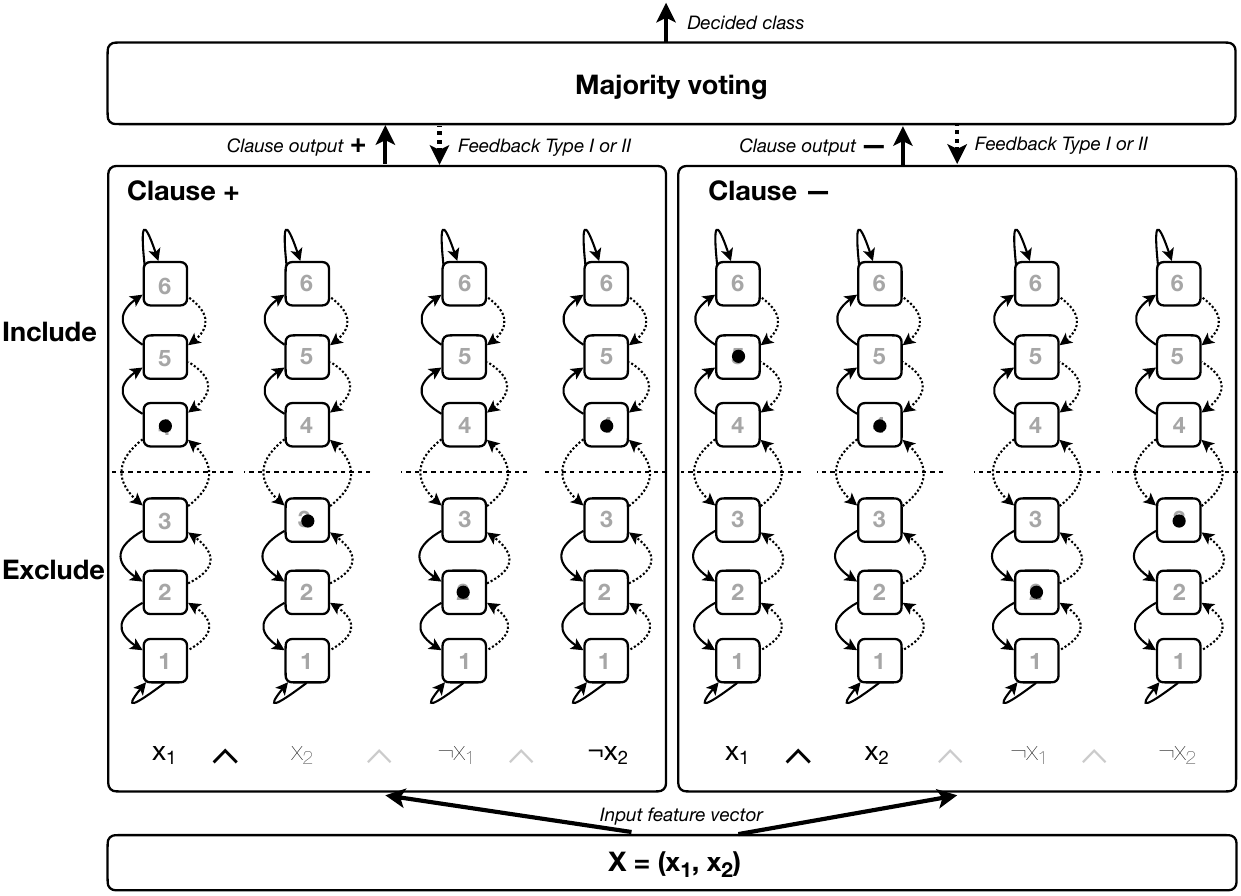}
\caption{Two Tsetlin Automata teams, each producing a conjunctive clause. The overall output is based on majority voting. }\label{figure:tm_architecture}
\end{figure}

Recall that a Tsetlin Automaton can be represented as a quintuple $\{\underline{\Phi}, \underline{\alpha}, \underline{\beta}, F(\cdot,\cdot), G(\cdot)\}$. Each clause $C^{\omega}_j, \omega \in \{0,1\}, j \in \{1, \ldots, n/2\}$, of a Tsetlin Machine is composed by a team of such Tsetlin Automata.
The team consists of $2o$ automata, one assigned per literal $l_k \in L$.  Let $\Phi^{\omega}_{jk} \in \{\phi_1, \phi_2 \ldots, \phi_{2N}\}$ refer to the state of the Tsetlin Automaton in  charge of literal $l_k$ for clause $C^{\omega}_j, \omega \in \{0,1\}, j \in \{1, \ldots, n/2\}$. It is this state that decides whether the literal is excluded (action $\alpha_1$) or included (action $\alpha_2$) in the clause.

Figure \ref{figure:tm_architecture} depicts two clauses, one with positive polarity ($C^1$) and one with negative polarity ($C^0$). They are both assigned four Tsetlin Automata, each with six states, controlling the inclusion of literals $L=\{x_1, x_2, \lnot x_1, \lnot x_2\}$. The leftmost Tsetlin Automaton in the figure has for example decided to include literal $x_1$ because it is in state $\phi_4$. The second Tsetlin Automaton, being in state $\phi_3$, has decided to exclude literal $x_2$. Given the decisions of the four  Tsetlin Automata of $C^1$, the clause has the form $x_1 \bar{x}_2$. Similarly, clause $C^0$ obtains the form $x_1 x_2$.

\subsection{The Tsetlin Machine Game for Learning Conjunctive Clauses}
\label{sec:tsetlin_machine_game}

We here introduce the game theoretic learning mechanism that guides the Tsetlin Automata towards solving the pattern recognition problem from Definition \ref{def:problem_definition}. The game is designed to deal with the problem of vanishing signal-to-noise ratio for learning automata games (cf. Section \ref{sec:vanishing}).

\subsubsection{Tsetlin Automata Games}

A game of Tsetlin Automata involves multiple automata and is played over several rounds~\cite{Narendra1989}. In each round of the game, the participating Tsetlin Automata independently decide upon their next action from $\underline{\alpha}$. Thus, if two actions are available to each automaton, they jointly select among $2^W$ unique action configurations, with $W$ being the number of automata. After the Tsetlin Automata have made their decisions, the round ends with the automata being individually penalized/rewarded based on the jointly selected action configuration. The next round of the game then starts.

To fully specify the game, we need to assign a reward probability per Tsetlin Automaton for each unique configuration of actions. We refer to these reward probabilities as the \emph{payoff matrix} of the game. With $W$ two-action automata, we accordingly need $W 2^W$ reward probabilities for the payoff matrix.

\subsubsection{Design of the Payoff Matrix}
The potential complexity of the Tsetlin Machine game is immense, because the decisions of every single Tsetlin Automaton jointly decide the behaviour of the Tsetlin Machine as a whole. Indeed, under the right conditions, a single Tsetlin Automaton has the power to completely disrupt a clause by introducing a contradiction. The payoffs of the game must therefore be carefully designed so that the Tsetlin Automata are guided towards building the clauses that solve the pattern recognition problem at hand. To complicate further, an explicit enumeration of the payoffs is impractical due to the potentially tremendous size of the payoff matrix.

To address the above challenges, we decompose the payoff matrix based on the notion of true positive, false negative, and false positive clause output, treating clauses as individual classifiers. True negative clause output is ignored to increase the freedom of the automata. By progressively suppressing false negative and false positive clause output, and reinforcing true positive clause output, we intend to guide the Tsetlin Automata towards maximizing pattern recognition accuracy. This guiding is based on what we will refer to as Type I and Type II Feedback. Based on the principles exposed in this section, we will define the resulting game formally in Section \ref{sec:theoretical_results}.

\subsubsection{Type I Feedback -- Combating False Negative Output}\label{sec:type_i_feedback}

In each round of the game, the Tsetlin Machine receives an input-output sample $(X,y)$. Type~I Feedback is then given to the Tsetlin Automata of clauses $C^{\omega}_j$ when $y=\omega$. Consider a particular Tsetlin Automaton with state $\Phi^{\omega}_{jk}$, controlling inclusion of literal $l_k$ in clause $C^{\omega}_j$, i.e., the $j$th clause of polarity $\omega$. Feedback is then decided by three factors:
\begin{enumerate}
    \item The decisions of the Tsetlin Automata team for clause $C_j^{\omega}(X)$ as a whole, manifested by the output of the clause.
    \item The truth value of the literal $l_k$ assigned to the automaton.
    \item The action $\alpha_z  = G(\Phi^{\omega}_{jk}), z \in \{1, 2\},$ decided by the automaton. 
\end{enumerate}
Table \ref{table:type_i_feedback} contains the probabilities that we use to generate  Type I Feedback, looking up the above three factors. For instance, assume that:
\begin{enumerate}
    \item Clause $C^{\omega}_j(X)$ evaluates to $1$,
    \item Literal $l_k$ is $1$, and
    \item Automaton state $\Phi^{\omega}_{jk}$ produces the \emph{Include} action ($\alpha_2$).
\end{enumerate}
By examining the corresponding cell in Table \ref{table:type_i_feedback}, we observe that the probability of receiving a reward $P(\mathrm{Reward})$ is $\frac{s-1}{s}$. Further, the probability of inaction $P(\mathrm{Inaction})$ is $\frac{1}{s}$. Finally, the probability of receiving a penalty $P(\mathrm{Penalty})$ is zero.

Note that the Inaction feedback is a novel extension to the Tsetlin Automaton, which traditionally receives either a Reward or a Penalty. When receiving the Inaction feedback, the Tsetlin Automaton is simply left unchanged.

\begin{table}[bh!]
\centering
\begin{tabular}{c|ccccc}
\multicolumn{2}{r|}{{\it Truth Value of Clause} $C_j^{\omega}$ }&\multicolumn{2}{c}{1}&\multicolumn{2}{c}{0}\\  
\multicolumn{2}{r|}{{\it Truth Value of Literal} $l_k$}&{1}&{0}&{1}&{0}\\
 \hline
 \hline
    \multirow{3}{*}{\bf Include Literal ($\alpha_2 \rightarrow l_k \in L_j^\omega$)}&\multicolumn{1}{c|}{$P(\mathrm{Reward})$}&$\frac{s-1}{s}$&NA&$0$&$0$\\
    &\multicolumn{1}{c|}{$P(\mathrm{Inaction})$}&$\frac{1}{s}$&NA&$\frac{s-1}{s}$&$\frac{s-1}{s}$\\
  &\multicolumn{1}{c|}{$P(\mathrm{Penalty})$}&$0$&NA&$\frac{1}{s} $&$\frac{1}{s}$\\
  \hline
  \multirow{3}{*}{\bf Exclude Literal ($\alpha_1 \rightarrow l_k \notin L_j^\omega$)}&\multicolumn{1}{c|}{$P(\mathrm{Reward})$}&$0$&$\frac{1}{s}$&$\frac{1}{s}$ &$\frac{1}{s}$\\
  &\multicolumn{1}{c|}{$P(\mathrm{Inaction})$}&$\frac{1}{s}$&$\frac{s-1}{s}$&$\frac{s-1}{s}$ &$\frac{s-1}{s}$\\
  &\multicolumn{1}{c|}{$P(\mathrm{Penalty})$}&$\frac{s-1}{s}$&$0$&$0$&$0$\\
  \hline
\end{tabular}
\caption{Type I Feedback --- Feedback from the perspective of a single Tsetlin Automaton deciding to either \emph{Include} or \emph{Exclude} a given literal $l_k$ in the clause $C^{\omega}_j$. Type I Feedback is triggered to increase the number of clauses that correctly evaluates to $1$ for a given input $X$.}
\label{table:type_i_feedback}
\end{table}

{\bf Brief Analysis of Type I Feedback.} Notice how the Type I Feedback probabilities force production of clauses that contain many literals, up to a certain point. We achieve this by reinforcing \emph{Include} actions more strongly than \emph{Exclude} actions, using reward- and penalty probabilities favouring \emph{Include}. As seen, the hyper-parameter $s, s \ge 1$, controls how strongly we favour \emph{Include}. In effect, $s$ decides how "fine-grained" patterns the clauses are going to capture. The larger the value of $s$, the more the Tsetlin Automata are stimulated to include literals in their clauses. The countering force is non-matching input, i.e., input that makes the clause evaluate to $0$. For such input, the \emph{Exclude} action is rewarded  and the \emph{Include} action is penalized. Clearly, the probability of encountering non-matching input grows as $s$ increases (more literals are included). When the above two counter-acting forces are in balance, we have a Nash equilibrium as discussed further in Section~\ref{sec:theoretical_results}. These dynamics are a critical part of the Tsetlin Machine, allowing learning of any sub-pattern, no matter how infrequent, as decided by~$s$.

{\bf Boosting of True Positive Feedback (Column 1 in Table \ref{table:type_i_feedback}).} The feedback probabilities in Table \ref{table:type_i_feedback} have been selected based on mathematical derivations (see Section \ref{sec:theoretical_results}). For certain real-life data sets, however, it turns out that further boosting of \emph{Include} actions can be beneficial. That is, pattern recognition accuracy can be enhanced by boosting rewarding of these actions when they produce true positive outcomes. Penalizing of \emph{Exclude} actions is then adjusted accordingly.  In all brevity, we boost rewarding in this manner by replacing $\frac{s-1}{s}$ with $1.0$ and $\frac{1}{s}$ with $0.0$ in Column 1 of Table \ref{table:type_i_feedback}.

\subsubsection{Type II Feedback -- Combating False Positive Output }

Table \ref{table:type_ii_feedback} covers Type II Feedback, that is, feedback that combats false positive output. Type~II Feedback is given to the Tsetlin Automata of clauses $C^{\omega}_j$ when $y\ne\omega$. The purpose of this feedback is to introduce candidate literals that increase discrimination power. That is, a clause $C^{\omega}_j$ is supposed to output $0$ when $y\ne\omega$, to discriminate between the outputs $\omega \in \{0, 1\}$.  However, if it erroneously evaluates to $1$, we correct this by identifying the Tsetlin Automata that have excluded a zero-valued literal from the clause. It is sufficient to merely include one of these literals to make the offending clause evaluate to $0$ instead. This is because the clauses are conjunctive.

{\bf Brief Analysis of Type II Feedback.} Observe that Type II Feedback only penalizes the \emph{exclusion} of literals of value $0$ when a clause outputs $1$ (false positive output). When a clause outputs $0$ (the two leftmost columns in the table), only Inaction-feedback is given.   This is to avoid local minima by leaving it to Type I feedback to reinforce \emph{Include} actions. Accordingly, together, Type I Feedback and Type II Feedback interact to minimize the expected output error, moving towards a global optimum as explored in Section~\ref{sec:theoretical_results}.

\begin{table}[bh!]
\centering
\begin{tabular}{c|ccccc}
\multicolumn{2}{r|}{\it Truth Value of Clause $C^{\omega}_j$}&\multicolumn{2}{c}{1}&\multicolumn{2}{c}{0}\\  
\multicolumn{2}{r|}{\it Truth Value of Literal $l_k$}&{1}&{0}&{1}&{0}\\
 \hline
 \hline
    \multirow{3}{*}{\bf Include Literal ($\alpha_2 \rightarrow l_k \in L_j^\omega$)}&\multicolumn{1}{c|}{$P(\mathrm{Reward})$}&$0$&$\mathrm{NA}$&$0$&$0$\\
    &\multicolumn{1}{c|}{$P(\mathrm{Inaction})$}&$1.0$&$\mathrm{NA}$&$1.0$&$1.0$\\
  &\multicolumn{1}{c|}{$P(\mathrm{Penalty})$}&$0$&$\mathrm{NA}$&$0$&$0$\\
  \hline
  \multirow{3}{*}{\bf Exclude Literal ($\alpha_1 \rightarrow l_k \notin L_j^\omega$)}&\multicolumn{1}{c|}{$P(\mathrm{Reward})$}&$0$&$0$&$0$&$0$\\
  &\multicolumn{1}{c|}{$P(\mathrm{Inaction})$}&$1.0$&$0$&$1.0$ &$1.0$\\
  &\multicolumn{1}{c|}{$P(\mathrm{Penalty})$}&$0$&$1.0$&$0$&$0$\\
  \hline
\end{tabular}
\caption{Type II Feedback --- Feedback from the perspective of a single Tsetlin Automaton deciding to either \emph{Include} or \emph{Exclude} a given literal $l_k$ in the clause $C^{\omega}_j$. Type II Feedback is triggered to increase the discrimination power of the clauses.}
\label{table:type_ii_feedback}
\end{table}

\subsubsection{Allocating Clauses to Sub-Patterns}

For the Tsetlin Machine output to be correct it is sufficient that the summation part of Eqn.~\ref{eqn:output_function}, $\sum_{j=1}^{n/2} C_j^{1}(X) -\sum_{j=1}^{n/2} C_j^{0}(X)$, has the correct sign. That is, it is sufficient that a single clause, with the appropriate polarity, outputs $1$. In the case of noisy data, however, it can be beneficial to introduce a margin $T$, referred to as the target of the summation. The intent is to make the available clauses distribute themselves across the sub-patterns in the data, so that $T$ clauses of the correct polarity evaluate to $1$ for any particular input $X$. Additionally, the summation target $T$ opens up for interplay between the clauses, including rectification of special cases.

The resource allocation mechanism that we introduce here is inspired by a finite state automaton-based solution to the fractional knapsack problem in unknown and stochastic environments \cite{Granmo2007d}. The purpose of the mechanism is to ensure that the Tsetlin Machine only spends a few of the available clauses to represent each specific sub-pattern. This is achieved by randomly selecting clauses to receive feedback, reducing intensity when approaching the summation target $T$.

{\bf Generating Type I Feedback.} For Type I Feedback, we randomly select clauses $C^{\omega}_j, \omega=y$. Among these, the probability of receiving Type I Feedback is
\begin{equation}
    \frac{T - \mathbf{clip}\left(\sum_{j=1}^{n/2} C_j^{1}(X) -\sum_{j=1}^{n/2} C_j^{0}(X), -T, T\right)}{2T}\label{eqn:activation_type_i}
\end{equation}
when $y=1$, and 
\begin{equation}
    \frac{T + \mathbf{clip}\left(\sum_{j=1}^{n/2} C_j^{1}(X) -\sum_{j=1}^{n/2} C_j^{0}(X), -T, T\right)}{2T}\label{eqn:activation_type_ib}
\end{equation}
when $y=0$. Here, the clip operation restricts the sum to lie between $-T$ and $T$.

{\bf Generating Type II Feedback.}  
For Type II Feedback, we randomly select clauses $C^{\omega}_j, \omega \ne y$. The probability of receiving Type II Feedback is:
\begin{equation}
    \frac{T + \mathbf{clip}\left(\sum_{j=1}^{n/2} C_j^{1}(X) -\sum_{j=1}^{n/2} C_j^{0}(X), -T, T\right)}{2T}\label{eqn:activation_type_ii}
\end{equation}
when $y=1$, and 
\begin{equation}
    \frac{T - \mathbf{clip}\left(\sum_{j=1}^{n/2} C_j^{1}(X) -\sum_{j=1}^{n/2} C_j^{0}(X), -T, T\right)}{2T}\label{eqn:activation_type_iib}
\end{equation}
when $y=0$.

Notice how the feedback vanishes as the number of triggering clauses correctly approaches $T$/$-T$. This is pertinent for effective use of the available pattern representation capacity. To exemplify, assume that the correct output is $y = 1$ for an input $X$. If the clause output sum accumulates to a total of $T$ or more, neither rewards nor penalties are provided to the involved Tsetlin Automata. This leaves the Tsetlin Automata free to learn other sub-patterns.

\subsubsection{The Tsetlin Machine Algorithm}

\begin{algorithm}
\caption{Tsetlin Machine}
\label{alg:tsetlin_machine}

\Input{Training data $(X, y) \in \mathcal{S} \sim P(X, y)$, Number of clauses $n$, Number of inputs $o$, Precision $s$, Target $T$}

\Output{Trained conjunctive clauses $C^{1}_j, C^{0}_j, j = 1, \ldots, n/2$} 

\begin{algorithmic} [1]
\Function{TrainTsetlinMachine}{$\mathcal{S},n,o,s,T$}

\State $\mathcal{A}^{1}_1, \ldots, \mathcal{A}^{1}_{n/2}  \gets$ CreateTATeams($n/2$,$2o$, $\phi_N$)\Comment{\parbox[t]{.4\linewidth}{Create $n/2$ teams of $2o$ Tsetlin Automata (TA), one team $\mathcal{A}^{1}_j$ per clause $C^{1}_j$. Each TA is initialized to state $\phi_N$.}}
\State $\mathcal{A}^{0}_1, \ldots, \mathcal{A}^{0}_{n/2}  \gets$ CreateTATeams($n/2$,$2o$, $\phi_N$)\Comment{\parbox[t]{.4\linewidth}{Create another $n/2$ teams of $2o$ TA, one team $\mathcal{A}^{0}_j$ per clause $C^{0}_j$. Each TA is initialized to state $\phi_N$.}}
\Repeat
    \State $X, y \gets$ GetTrainingExample($\mathcal{S}$)\Comment{Mini-batches, random selection, etc.}
    \State $C^{1}_1, \ldots, C^{1}_{n/2} \gets$ ObtainClauses($\mathcal{A}^{1}_1, \ldots, \mathcal{A}^{1}_{n/2}$) \Comment{\parbox[t]{.35\linewidth}{Each TA team $\mathcal{A}^{1}_j$ produces a clause $C^{1}_j$.}}
    \State $C^{0}_1, \ldots, C^{0}_{n/2} \gets$ ObtainClauses($\mathcal{A}^{0}_1, \ldots, \mathcal{A}^{0}_{n/2}$)\Comment{\parbox[t]{.35\linewidth}{Each TA team $\mathcal{A}^{0}_j$ produces a clause $C^{0}_j$.}}
    \State $v \gets \sum_{j=1}^{n/2} C_j^{1}(X) -\sum_{j=1}^{n/2} C_j^{0}(X)$ \Comment{Calculate sum of clause outputs.}
    \For{$j \gets 1,\ldots, n/2$} \Comment{Give feedback to the TA teams.}
        \If{$y = 1$}
            \If{Random() $\le \frac{T - \mathbf{clip}\left(v, -T, T\right)}{2T}$}
                 \State GenerateTypeIFeedback($X,  C^{1}_j(X),  \mathcal{A}^{1}_j, o$) \Comment{\parbox[t]{.275\linewidth}{Give Type I Feedback to TA of clause $C^{1}_j$.}}
            \EndIf
            \If{Random() $\le \frac{T - \mathbf{clip}\left(v, -T, T\right)}{2T}$}
                \State GenerateTypeIIFeedback($X, C^{0}_j(X), \mathcal{A}^{0}_j, o$) 
                \Comment{\parbox[t]{.275\linewidth}{Give Type II Feedback to TA of clause $C^{0}_j$.}}
            \EndIf
        \ElsIf{$y = 0$}
            \If{Random() $\le \frac{T + \mathbf{clip}\left(v, -T, T\right)}{2T}$}
                \State GenerateTypeIIFeedback($X, C^{1}_j(X),\mathcal{A}^{1}_j, o$) 
                \Comment{\parbox[t]{.275\linewidth}{Give Type II Feedback to TA of clause $C^{1}_j$.}}
            \EndIf
            \If{Random() $\le \frac{T + \mathbf{clip}\left(v, -T, T\right)}{2T}$}
                 \State GenerateTypeIFeedback($X, C^{0}_j(X),\mathcal{A}^{0}_j, o$) \Comment{\parbox[t]{.275\linewidth}{Give Type I Feedback to TA of clause $C^{0}_j$.}}
            \EndIf
        \EndIf
    \EndFor
\Until{StopCriteria($\mathcal{S}, C^{0}_1, \ldots, C^{0}_{n/2}, C^{1}_1, \ldots, C^{1}_{n/2}$)}
\State \Return PruneAllExcludeClauses($C^{0}_1, \ldots, C^{0}_{n/2}, C^{1}_1, \ldots, C^{1}_{n/2}$) \Comment{Return completely trained conjunctive clauses after pruning clauses where all literals have been excluded.}
\EndFunction
\end{algorithmic}
\end{algorithm}

The step-by-step procedure for learning conjunctive clauses can be found in Algorithm \ref{alg:tsetlin_machine}. The algorithm takes a set of training examples $(X, y) \in \mathcal{S}$ as input. It then produces a collection of conjunctive clauses for predicting the output $y$. We will now take a closer look at the algorithm, line-by-line.

{\bf Lines 2-3.} From the perspective of game theory, we form one team of Tsetlin Automata per clause, denoted $\mathcal{A}^{0}_j$, $\mathcal{A}^{1}_j$, $j \in \{1, \ldots, n/2\}$. A team consists of $2o$ automata, each assigned a user specified number of states, $\Phi^{\omega}_{jk} \in \{\phi_1,\phi_2, \ldots, \phi_{2N}\}$, $j \in \{1, \ldots, n/2\}$, $k \in \{1,\ldots, 2o\}$, $\omega \in \{0,1\}$. The initial state is then set to 
$\Phi^{\omega}_{jk} \gets \phi_N$ (the \emph{Exclude} action).

{\bf Line 5.} The learning process is driven by a set of training examples $\mathcal{S}$, sampled from the input-output distribution $P(X, y)$. Each training example $(X, y)$ is fed to the Tsetlin Machine, one at a time, facilitating online learning.

{\bf Line 6-7.}  In each iteration, the Tsetlin Automata decide whether to \emph{Include} or \emph{Exclude} their assigned literals in their respective clauses. The result is a new  set of clauses, $C^{0}_1, \ldots, C^{0}_{n/2},$ $C^{1}_1, \ldots, C^{1}_{n/2}$, for predicting $y$.

{\bf Line 8.} Based on the new clauses, the current example $X$ is evaluated, producing a clause sum $v$ according to Eqn. \ref{eqn:output_function}. The unit step function $u$ is not used here because examples are not classified during training.

{\bf Lines 9-25.} The next step is to randomly determine the clauses whose Tsetlin Automata are to receive Type I Feedback (Eqn. \ref{eqn:activation_type_i}-\ref{eqn:activation_type_ib}) and Type II Feedback (Eqn. \ref{eqn:activation_type_ii}-\ref{eqn:activation_type_iib}), based on the clipped difference between $v$ and $T$.

After Type I or Type II Feedback have been triggered for a clause, the Tsetlin Automata of the clause are rewarded/penalized according to Algorithm \ref{alg:type_i_feedback} and Algorithm \ref{alg:type_ii_feedback}. In all brevity, rewarding/penalizing is directly based on Table \ref{table:type_i_feedback} and Table \ref{table:type_ii_feedback}.

\begin{algorithm}[t]
\caption{Type I Feedback}
\label{alg:type_i_feedback}

\Input{Input vector $X$ of size $o$, Clause output $c^{\omega}_j$, Tsetlin Automata team $\mathcal{A}^{\omega}_j$}

\begin{algorithmic} [1]
\Procedure{GenerateTypeIFeedback}{$X, c^{\omega}_j, \mathcal{A}^{\omega}_j, o$}

\For{$k \gets 1, \ldots , 2o$}\Comment{Reward/Penalize all Tsetlin Automata in $\mathcal{A}^{\omega}_j$.}
    \State $l_k \gets \mathrm{ObtainLiteral}(X, k)$ \Comment{Get literal $l_k$ from input $X$.}
    \State $\alpha_u \gets G(\Phi^{\omega}_{jk})$ \Comment{Obtain corresponding automaton action according to Eqn. \ref{eqn:action_mapping}.}
    \State $\beta_v \gets$
    SampleTypeIFeedback($\alpha_u, l_k, c^{\omega}_j$) \Comment{Sample Type I Feedback from Table \ref{table:type_i_feedback}.}
    \State $\Phi^{\omega}_{jk} \gets F(\Phi^{\omega}_{jk}, \beta_v)$ \Comment{Update state of Tsetlin Automaton according to Eqn. \ref{eqn:state_transition}.}
\EndFor

\EndProcedure
\end{algorithmic}
\end{algorithm}

\begin{algorithm}[t]
\caption{Type II Feedback}
\label{alg:type_ii_feedback}

\Input{Input vector $X$ of size $o$, Clause output $c^{\omega}_j$, Tsetlin Automata team $\mathcal{A}^{\omega}_j$}

\begin{algorithmic} [1]
\Procedure{GenerateTypeIIFeedback}{$X, c^{\omega}_j, \mathcal{A}^{\omega}_j, o$}

\For{$k \gets 1, \ldots , 2o$}\Comment{Penalize all Tsetlin Automata in $\mathcal{A}^{\omega}_j$.}
    \State $l_k \gets \mathrm{ObtainLiteral}(X, k)$ \Comment{Get literal $l_k$ from input $X$.}
    \State $\alpha_u \gets G(\Phi^{\omega}_{jk})$ \Comment{Obtain corresponding automaton action according to Eqn. \ref{eqn:action_mapping}.}
    \State $\beta_v \gets$
    SampleTypeIIFeedback($\alpha_u, l_k, c^{\omega}_j$) \Comment{Sample Type II Feedback from Table \ref{table:type_ii_feedback}.}
    \State $\Phi^{\omega}_{jk} \gets F(\Phi^{\omega}_{jk}, \beta_v)$ \Comment{Update state of Tsetlin Automaton according to Eqn. \ref{eqn:state_transition}.}
\EndFor

\EndProcedure
\end{algorithmic}
\end{algorithm}

{\bf Line 26.} The above steps are iterated until a stopping criteria is fulfilled (for instance a certain number of iterations over the dataset). The current clauses are then returned as the output of the learning process, after the clauses without literals have been removed.

\subsection{Implementation Using Bitwise Operators}

Small memory footprint and speed of operation can be crucial in complex and large scale pattern recognition. Being based on propositional formula, the Tsetlin Machine architecture can naturally be represented with bits and manipulated upon using bitwise operators. However, it is not straightforward how to represent and update the Tsetlin Automata themselves. First of all, the state index of each Tsetlin Automaton is an integer value. Further, the action of an automaton is decided upon using a smaller-than operator, while feedback is processed by means of increment and decrement operations.

\begin{figure}[!th]
\centering
\includegraphics[width=3.0in]{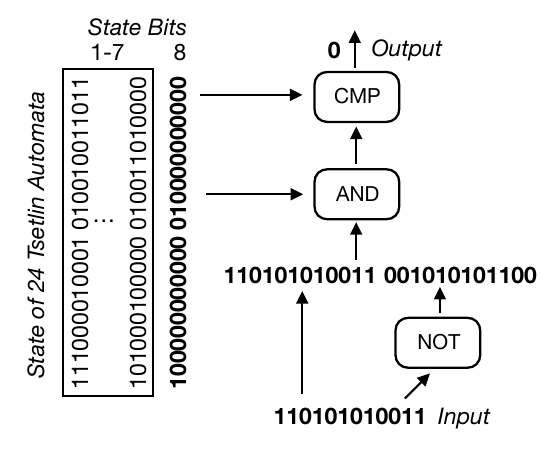}
\caption{Bit-based representation of a clause for the Noisy XOR dataset (see Section \ref{sec:empirical_results}). The bit-based representation of the Tsetlin Automata states allows the actions of all of the automata to be obtained directly from the most significant bit (bit 8 in the figure).}
\label{figure:bit_based_manipulation}
\end{figure}

One approach to bitwise operation is to jointly represent the state indexes of all of the Tsetlin Automata of a clause $C_j^\omega$ with multiple sequences of bits. Sequence 1 then contains the first bit of each state index, sequence 2 contains the second bit, and so on, as exemplified in Figure \ref{figure:bit_based_manipulation} for 24 Tsetlin Automata. The benefit of this representation is that the action of each Tsetlin Automaton is readily available from the most significant bit (sequence 8 in the figure). Thus, the output of the clause can be obtained from the input based on fast bitwise operators (NOT, AND, and CMP - comparison).

In the figure, a setup for the Noisy XOR dataset from Section \ref{sec:empirical_results} is used for illustration purposes. First, the 12-bit input is extended to 24 bits by concatenating the original input with the original input inverted. The resulting 24 bits are in turn connected with 24 Tsetlin Automata. The first 12 control the non-negated input, while the second 12 control the negated input.

Employing the latter bit-based representation reduces memory usage four times, compared to using a full 32-bit integer to represent the state of each and every Tsetlin Automaton. More importantly, it is possible to increment/decrement the states of all of the automata in parallel with bitwise operations through customized increment/decrement procedures, significantly increasing learning speed. As an example, for the MNIST dataset (cf. Section \ref{sec:empirical_results}), the overall memory usage is approximately ten times smaller, learning speed 3.5 times faster, and classification speed 8 times faster with the bit-based representation.

When deployed after training, only the state bit sequence containing the most significant bit is required. The other bit sequences can be discarded because these bits are only used to keep track of the learning. This provides a further reduction in memory usage.

\section{Theoretical Analysis}
\label{sec:theoretical_results}

In this section, we analyse the Tsetlin Machine game formally. We base the analysis on partitioning the input space $\mathcal{X}$ into subsets that isolate the impact a single Tsetlin Automaton has on classification, mediated through its clause. We then study each of these subsets for the two output scenarios $y=0$ and $y=1$. From this investigation, we identify the Nash equilibria of Tsetlin Machine learning and compare the Nash equilibria clauses with the optimal sub-patterns from Definition \ref{def:problem_definition}.

\subsection{The Payoff Matrix of the Tsetlin Machine Game}

Based on the learning dynamics described algorithmically in Section \ref{sec:tsetlin_machine}, we now introduce a more succinct description of Tsetlin Machine learning by formally defining the payoff matrix of the Tsetlin Machine game.
\begin{definition}\label{def:tsetlin_machine_game}
Consider a pattern recognition problem per Definition \ref{def:problem_definition} and a Tsetlin Machine with $n$ clauses $C_j^\omega(X), j \in \{1,\ldots, n/2\}, \omega \in \{0, 1\}$. The resulting game consists of $2no$ Tsetlin Automata (players), one automaton per literal $l_k \in L = \{x_1,\ldots,x_o, \bar{x}_1,\ldots, \bar{x}_o\}$, per clause $C_j^\omega(X)$. Let $A^\omega_{jk}$ be a Tsetlin Automaton with actions $a^\omega_{jk} \in \{\alpha_1, \alpha_2\}$, controlling the inclusion of literal $l_k$ in clause $C_j^\omega$ ($\alpha_1$ represents Exclude and $\alpha_2$ represents Include). 
Each cell in the payoff matrix of the game then refers to a unique action configuration $\overline{\alpha} = (a^0_{1,1}, \ldots, a^0_{n/2,2o}, a^1_{1,1} \ldots, a^1_{n/2,2o}) \in \{\alpha_1, \alpha_2\}^{2no}$. Accordingly, the payoff matrix consists of $2^{2no}$ cells. For every cell, we define the payoff of each $a^\omega_{jk}$-action in $\overline{\alpha}$ using the four stochastic variables $X, y, U^\omega_j$, and $R^\omega_{jk}$:
\begin{itemize}
    \item Each game round, the distribution $P(X, y)$ randomly produces a training example $(X, y)$.
    \item Let the binary stochastic variable $U^\omega_j$ refer to whether clause $C^\omega_j$ is updated ($U^\omega_j=1$) or ignored ($U^\omega_j=0$) in the given game round. The distribution of $U^\omega_j$ is decided as follows. First, the clipped clause sum $v^c = \mathbf{clip}\left(\sum_{j=1}^{n/2} C_j^1(X) -\sum_{j=1}^{n/2} C_j^0(X), -T, T\right)$ is determined. This clipped clause sum is calculated from the clauses produced by the joint action configuration $\overline{\alpha}$ of the payoff matrix cell, evaluated upon $X$. Clause $C_j^\omega$ is then updated with probability $P(U^\omega_j=1) = \frac{T-v^c}{2T}$ if $y=1$  and with probability $P(U^\omega_j=1) = \frac{T+v^c}{2T}$ if $y=0$.
    \item Finally, we decide the payoff $R^\omega_{jk} \in \{-1, 0, +1\}$ (Penalty, Inaction, Reward) for each automaton $A^\omega_{jk}$. If $U^\omega_j=0$ the payoff is $0$. Otherwise, $R^\omega_{jk}$ is randomly set based on Table~\ref{table:type_i_feedback} if $y = \omega$ (Type I Feedback) and Table~\ref{table:type_ii_feedback} if $y \neq \omega$ (Type II Feedback). We obtain the distribution of $R^\omega_{jk}$ from the designated table by looking up: (1) the action selected ($\alpha_1$ or $\alpha_2$), (2) the value of literal $l_k$ from $X$, and (3) the value of the clause $C^\omega_j(X)$.
\end{itemize}
Accordingly, the payoff of each action $a^\omega_{jk}$ in $\overline{\alpha}$ becomes $U^\omega_j \cdot R^\omega_{jk}$.
\end{definition}
In the following, we analyze the dynamics of the game using the above payoff matrix.

\begin{figure}[!th]
\centering
\includegraphics[width=6.0in]{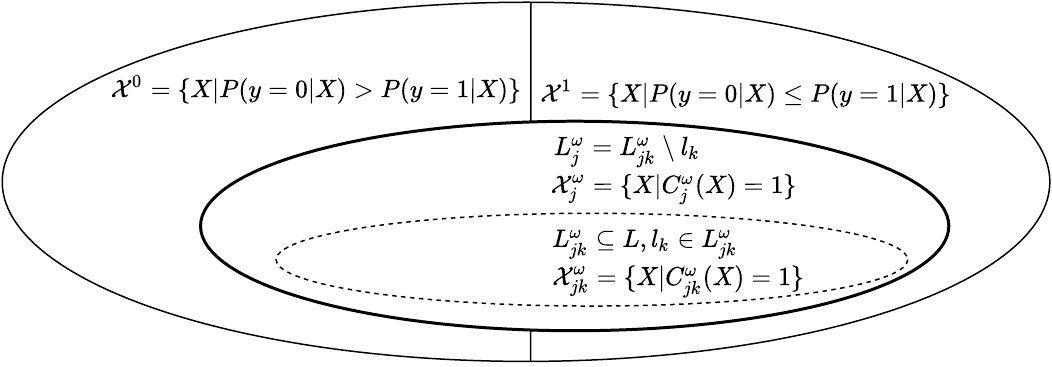}
\caption{The subsets $\mathcal{X}^0$, $\mathcal{X}^1$, $\mathcal{X}_j^\omega$, and $\mathcal{X}_{jk}^\omega$, pertinent for the theoretical analysis.}
\label{figure:pertinent_subsets}
\end{figure}

\subsection{Pertinent Input Space Subsets}
The input space subsets ($\mathcal{X}$-subsets) of interest are listed in Table \ref{table:expected_feedback_exclude} and Table \ref{table:expected_feedback_include}. Observe that all of the $\mathcal{X}$-subsets in the tables are derived from the following four subsets: $\mathcal{X}^0$, $\mathcal{X}^1$, $\mathcal{X}_j^1$, and $\mathcal{X}_{jk}^1$, illustrated in Figure \ref{figure:pertinent_subsets}.

The subset $\mathcal{X}^1=\{X | P(y = 0 | X) \le
P(y = 1 | X)\}$ is the subset of the input space where it is optimal to output $y=1$. Conversely, the subset $\mathcal{X}^0=\{X | P(y = 0 | X) >
P(y = 1 | X)\}$ is the compliment of $\mathcal{X}$, where it is optimal to output $y=0$. The subset 
$\mathcal{X}_{jk}^1 = \{X | C_{jk}^\omega(X) = 1\}$, $L_{jk}^\omega \subseteq L, l_k \in L_{jk}$, is the part of the input space in which the clause $C_{jk}^\omega(X)$ with literals $L_{jk}^\omega, l_k \in L_{jk}^\omega,$ evaluates to $1$. Finally, the subset 
$\mathcal{X}_j^1 = \{X | C_j^\omega(X) = 1\}, L_j^\omega= L_{jk}^\omega \setminus l_k$, is a superset of  $\mathcal{X}_{jk}^1$, where the literal $l_k$ has been removed from the conjunction.

Recall that it is a single Tsetlin Automaton $A^\omega_{jk}$ that decides whether to \emph{Include} or \emph{Exclude}  literal $l_k$ above, which means that it in effect decides between clause $C_{jk}^\omega(X)$ (literal included) and clause $C_j^\omega(X)$ (literal excluded).

Further note that we in this section ignore polarity to simplify notation. That is, we analyze how the clauses with positive polarity in the Tsetlin Machine game capture the sub-patterns $Q^1_h, h \in \{1, \ldots, n/2\},$ from Definition \ref{def:problem_definition}. The analysis for negative polarity clauses follows the same structure.

\begin{table}[!!b]
\centering
\begin{small}
\begin{tabular}{l|l|c|l|l} 
$\mathbf{y}$&$\mathcal{X}$-\textbf{Subset}&\textbf{R}&\textbf{Probability}&\textbf{Noise $1-\gamma$}\\
\hline
$0$&$\mathcal{X}_{jk}^1 \cap \mathcal{X}^1$&$0$&$P(y=0 |  \mathcal{X}_{jk}^1 \cap \mathcal{X}^1) P(\mathcal{X}_{jk}^1 \cap \mathcal{X}^1)$&$(1 - \gamma) P(\mathcal{X}_{jk}^1 \cap \mathcal{X}^1)$\\
$0$&$(\mathcal{X}_j^1 \setminus \mathcal{X}_{jk}^1) \cap \mathcal{X}^1 $&$-1$&$P(y=0 |  [\mathcal{X}_j^1 \setminus \mathcal{X}_{jk}^1] \cap \mathcal{X}^1) P([\mathcal{X}_j^1 \setminus \mathcal{X}_{jk}^1] \cap \mathcal{X}^1)$&$(1 - \gamma) P([\mathcal{X}_j^1 \setminus \mathcal{X}_{jk}^1] \cap \mathcal{X}^1)$\\
$0$&$\mathcal{X}^1 \setminus \mathcal{X}_j^1$&$0$&$P(y=0 | \mathcal{X}^1 \setminus \mathcal{X}_j^1) P(\mathcal{X}^1 \setminus \mathcal{X}_j^1)$&$(1 - \gamma) P(\mathcal{X}^1 \setminus \mathcal{X}_j^1)$\\
$0$&$\mathcal{X}_{jk}^1 \cap \mathcal{X}^0$&$0$&$P(y=0 |  \mathcal{X}_{jk}^1 \cap \mathcal{X}^0) P( \mathcal{X}_{jk}^1 \cap \mathcal{X}^0)$&$\gamma P( \mathcal{X}_{jk}^1 \cap \mathcal{X}^0)$\\
$0$&$(\mathcal{X}_j^1 \setminus \mathcal{X}_{jk}^1) \cap \mathcal{X}^0 $&$-1$&$P(y=0 | [\mathcal{X}_j^1 \setminus \mathcal{X}_{jk}^1] \cap \mathcal{X}^0) P([\mathcal{X}_j^1 \setminus \mathcal{X}_{jk}^1] \cap \mathcal{X}^0)$&$\gamma P([\mathcal{X}_j^1 \setminus \mathcal{X}_{jk}^1] \cap \mathcal{X}^0)$\\
$0$&$\mathcal{X}^0 \setminus \mathcal{X}_j^1$&$0$&$P(y=0 | \mathcal{X}^0 \setminus \mathcal{X}_j^1) P( \mathcal{X}^0 \setminus \mathcal{X}_j^1)$&$\gamma P( \mathcal{X}^0 \setminus \mathcal{X}_j^1)$\\
\hline
$1$&$\mathcal{X}_{jk}^1 \cap \mathcal{X}^1$&$-\frac{s-1}{s}$&$P(y=1 |  \mathcal{X}_{jk}^1 \cap \mathcal{X}^1) P(\mathcal{X}_{jk}^1 \cap \mathcal{X}^1)$&$\gamma P(\mathcal{X}_{jk}^1 \cap \mathcal{X}^1)$\\
$1$&$(\mathcal{X}_j^1 \setminus \mathcal{X}_{jk}^1) \cap \mathcal{X}^1$&$\frac{1}{s}$&$P(y=1 |  [\mathcal{X}_j^1 \setminus \mathcal{X}_{jk}^1] \cap \mathcal{X}^1) P([\mathcal{X}_j^1 \setminus \mathcal{X}_{jk}^1] \cap \mathcal{X}^1)$&$\gamma P([\mathcal{X}_j^1 \setminus \mathcal{X}_{jk}^1] \cap \mathcal{X}^1)$\\
$1$&$\mathcal{X}^1 \setminus \mathcal{X}_j^1$&$\frac{1}{s}$&$P(y=1 | \mathcal{X}^1 \setminus \mathcal{X}_j^1) P( \mathcal{X}^1 \setminus \mathcal{X}_j^1)$&$\gamma P(\mathcal{X}^1 \setminus \mathcal{X}_j^1)$\\
$1$&$\mathcal{X}_{jk}^1 \cap \mathcal{X}^0$&$-\frac{s-1}{s}$&$P(y=1 |  \mathcal{X}_{jk}^1 \cap \mathcal{X}^0) P( \mathcal{X}_{jk}^1 \cap \mathcal{X}^0)$&$(1-\gamma)P( \mathcal{X}_{jk}^1 \cap \mathcal{X}^0)$\\
$1$&$(\mathcal{X}_j^1 \setminus \mathcal{X}_{jk}^1) \cap \mathcal{X}^0 $&$\frac{1}{s}$&$P(y=1 |  [\mathcal{X}_j^1 \setminus \mathcal{X}_{jk}^1] \cap \mathcal{X}^0) P([\mathcal{X}_j^1 \setminus \mathcal{X}_{jk}^1] \cap \mathcal{X}^0)$&$(1-\gamma)P([\mathcal{X}_j^1 \setminus \mathcal{X}_{jk}^1] \cap \mathcal{X}^0)$\\
$1$&$\mathcal{X}^0 \setminus \mathcal{X}_j^1$&$\frac{1}{s}$&$P(y=1 | \mathcal{X}^0 \setminus \mathcal{X}_j^1) P( \mathcal{X}^0 \setminus \mathcal{X}_j^1)$&$(1-\gamma) P( \mathcal{X}^0 \setminus \mathcal{X}_j^1)$
\end{tabular}
\end{small}
\caption{Expected payoff for the \emph{Exclude} action.}\label{table:expected_feedback_exclude}
\end{table}

\begin{table}[!!b]
\centering
\begin{small}
\begin{tabular}{l|l|c|l|l} 
$\mathbf{y}$&$\mathcal{X}$-\textbf{Subset}&\textbf{R}&\textbf{Probability}&\textbf{Noise $1-\gamma$}\\
\hline
$0$&$\mathcal{X}_{jk}^1 \cap \mathcal{X}^1$&$0$&$P(y=0 |  \mathcal{X}_{jk}^1 \cap \mathcal{X}^1) P(\mathcal{X}_{jk}^1 \cap \mathcal{X}^1)$&$(1 - \gamma) P(\mathcal{X}_{jk}^1 \cap \mathcal{X}^1)$\\
$0$&$(\mathcal{X}_j^1 \setminus \mathcal{X}_{jk}^1) \cap \mathcal{X}^1 $&$0$&$P(y=0 |  [\mathcal{X}_j^1 \setminus \mathcal{X}_{jk}^1] \cap \mathcal{X}^1) P([\mathcal{X}_j^1 \setminus \mathcal{X}_{jk}^1] \cap \mathcal{X}^1)$&$(1 - \gamma) P([\mathcal{X}_j^1 \setminus \mathcal{X}_{jk}^1] \cap \mathcal{X}^1)$\\
$0$&$\mathcal{X}^1 \setminus \mathcal{X}_j^1$&$0$&$P(y=0 | \mathcal{X}^1 \setminus \mathcal{X}_j^1) P(\mathcal{X}^1 \setminus \mathcal{X}_j^1)$&$(1 - \gamma) P(\mathcal{X}^1 \setminus \mathcal{X}_j^1)$\\
$0$&$\mathcal{X}_{jk}^1 \cap \mathcal{X}^0$&$0$&$P(y=0 |  \mathcal{X}_{jk}^1 \cap \mathcal{X}^0) P( \mathcal{X}_{jk}^1 \cap \mathcal{X}^0)$&$\gamma P( \mathcal{X}_{jk}^1 \cap \mathcal{X}^0)$\\
$0$&$(\mathcal{X}_j^1 \setminus \mathcal{X}_{jk}^1) \cap \mathcal{X}^0 $&$0$&$P(y=0 | [\mathcal{X}_j^1 \setminus \mathcal{X}_{jk}^1] \cap \mathcal{X}^0) P([\mathcal{X}_j^1 \setminus \mathcal{X}_{jk}^1] \cap \mathcal{X}^0)$&$\gamma P([\mathcal{X}_j^1 \setminus \mathcal{X}_{jk}^1] \cap \mathcal{X}^0)$\\
$0$&$\mathcal{X}^0 \setminus \mathcal{X}_j^1$&$0$&$P(y=0 | \mathcal{X}^0 \setminus \mathcal{X}_j^1) P( \mathcal{X}^0 \setminus \mathcal{X}_j^1)$&$\gamma P( \mathcal{X}^0 \setminus \mathcal{X}_j^1)$\\
\hline
$1$&$\mathcal{X}_{jk}^1 \cap \mathcal{X}^1$&$\frac{s-1}{s}$&$P(y=1 |  \mathcal{X}_{jk}^1 \cap \mathcal{X}^1) P(\mathcal{X}_{jk}^1 \cap \mathcal{X}^1)$&$\gamma P(\mathcal{X}_{jk}^1 \cap \mathcal{X}^1)$\\
$1$&$(\mathcal{X}_j^1 \setminus \mathcal{X}_{jk}^1) \cap \mathcal{X}^1$&$-\frac{1}{s}$&$P(y=1 |  [\mathcal{X}_j^1 \setminus \mathcal{X}_{jk}^1] \cap \mathcal{X}^1) P([\mathcal{X}_j^1 \setminus \mathcal{X}_{jk}^1] \cap \mathcal{X}^1)$&$\gamma P([\mathcal{X}_j^1 \setminus \mathcal{X}_{jk}^1] \cap \mathcal{X}^1)$\\
$1$&$\mathcal{X}^1 \setminus \mathcal{X}_j^1$&$-\frac{1}{s}$&$P(y=1 | \mathcal{X}^1 \setminus \mathcal{X}_j^1) P( \mathcal{X}^1 \setminus \mathcal{X}_j^1)$&$\gamma P(\mathcal{X}^1 \setminus \mathcal{X}_j^1)$\\
$1$&$\mathcal{X}_{jk}^1 \cap \mathcal{X}^0$&$\frac{s-1}{s}$&$P(y=1 |  \mathcal{X}_{jk}^1 \cap \mathcal{X}^0) P( \mathcal{X}_{jk}^1 \cap \mathcal{X}^0)$&$(1-\gamma)P( \mathcal{X}_{jk}^1 \cap \mathcal{X}^0)$\\
$1$&$(\mathcal{X}_j^1 \setminus \mathcal{X}_{jk}^1) \cap \mathcal{X}^0 $&$-\frac{1}{s}$&$P(y=1 |  [\mathcal{X}_j^1 \setminus \mathcal{X}_{jk}^1] \cap \mathcal{X}^0) P([\mathcal{X}_j^1 \setminus \mathcal{X}_{jk}^1] \cap \mathcal{X}^0)$&$(1-\gamma)P([\mathcal{X}_j^1 \setminus \mathcal{X}_{jk}^1] \cap \mathcal{X}^0)$\\
$1$&$\mathcal{X}^0 \setminus \mathcal{X}_j^1$&$-\frac{1}{s}$&$P(y=1 | \mathcal{X}^0 \setminus \mathcal{X}_j^1) P( \mathcal{X}^0 \setminus \mathcal{X}_j^1)$&$(1-\gamma) P( \mathcal{X}^0 \setminus \mathcal{X}_j^1)$
\end{tabular}
\end{small}
\caption{Expected payoff for the \emph{Include} action.}\label{table:expected_feedback_include}
\end{table}

\subsection{Expected Payoff for the Actions of a Single Tsetlin Automaton}\label{sec:expected_feedback}

As defined in the payoff matrix of the Tsetlin Machine game, the stochasticity in Tsetlin Machine learning comes from: (i) the training samples $(X, y)$, which are drawn randomly according to an unknown distribution $P(X, y)$; (ii) the random selection of clauses for updating; and (iii) the random generation of the rewards and penalties from Type~I feedback (Table \ref{table:type_i_feedback}). \emph{We first focus on what happens with a specific Tsetlin Automaton each time its clause activates. Thus, for the moment, we ignore stochasticity stemming from random clause selection, studying a single clause.}

We have collected the information needed to calculate the expected payoff for the \emph{Exclude} action in Table \ref{table:expected_feedback_exclude} and for the \emph{Include} action in Table \ref{table:expected_feedback_include}. The first column of the table enumerates the possible $y$-values. The second column specifies a partitioning of the input space $\mathcal{X}$ for each $y$-value. The third column contains the expected payoff of the action covered by the table, for the given $\mathcal{X}$-subset and output $y$. The expected payoff has been obtained from Table~\ref{table:type_i_feedback} and Table~\ref{table:type_ii_feedback}. The fourth column states the probability of obtaining a training example $(X, y$) for the corresponding $\mathcal{X}$-subset and output $y$. The fifth column specifies these probabilities assuming a static output noise probability $P(y = 0| \mathcal{X}^1) = P(y = 1| \mathcal{X}^0) = 1 - \gamma, \gamma > 0.5$. If for instance $\gamma = 0.75$, the output is erroneously $y = 0$ for $X \in \mathcal{X}^1$ and $y = 1$ for $X \in \mathcal{X}^0$, on average, $25\%$ of the time. The assumption of static noise is not a limitation of the Tsetlin Machine, however, the assumption simplifies the theoretical analysis that follows.

From Table \ref{table:expected_feedback_exclude} and Table \ref{table:expected_feedback_include}, we can calculate the expected payoff of each action by summing up the rewards (positive polarity) and penalties (negative polarity) in the $\mathbf{R}$-column, each multiplied with the probability of the $\mathcal{X}$-subset and $y$-value of its row. The addend from the first row in Table \ref{table:expected_feedback_exclude} is for instance $0 \cdot P(y=0 |  \mathcal{X}_{jk}^1 \cap \mathcal{X}^1) P(\mathcal{X}_{jk}^1 \cap \mathcal{X}^1)$, which becomes zero because the reward is zero.

\begin{mylemma}\label{lemma:expected_reward_exclude}
Consider a literal $l_k$ and clause $C_j^\omega$. The expected payoff of excluding $l_k$ from $C_j^\omega$ in the Tsetlin Machine game (Definition \ref{def:tsetlin_machine_game}) is: 
\begin{eqnarray}
\frac{1}{s} \cdot \gamma \cdot P(\mathcal{X}^1 \setminus \mathcal{X}_{jk}^1) +
\frac{1}{s} \cdot (1-\gamma) \cdot  P(\mathcal{X}^0 \setminus \mathcal{X}_{jk}^1) -
\frac{s-1}{s} \cdot  \gamma \cdot  P(\mathcal{X}_{jk}^1 \cap \mathcal{X}^1)&-\nonumber\\
\frac{s-1}{s} \cdot (1-\gamma) \cdot  P(\mathcal{X}_{jk}^1 \cap \mathcal{X}^0) -
(1 - \gamma) \cdot  P([\mathcal{X}_j^1 \setminus \mathcal{X}_{jk}^1] \cap \mathcal{X}^1) -
\gamma \cdot P([\mathcal{X}_j^1 \setminus \mathcal{X}_{jk}^1] \cap \mathcal{X}^0).&\label{eqn:expected_reward_exclude}
\end{eqnarray}
\end{mylemma}

\begin{proof}
We prove Lemma \ref{lemma:expected_reward_exclude} based on Table  \ref{table:expected_feedback_exclude}.
The expected payoff of the \emph{Exclude} action can be calculated as follows, assuming static noise probability $1-\gamma$:
\begin{eqnarray}
\frac{1}{s} \cdot \gamma \cdot P([\mathcal{X}_j^1 \setminus \mathcal{X}_{jk}^1] \cap \mathcal{X}^1)&+\nonumber\\
\frac{1}{s} \cdot \gamma \cdot P(\mathcal{X}^1 \setminus \mathcal{X}_j^1)&+\nonumber\\
\frac{1}{s} \cdot (1-\gamma) \cdot P([\mathcal{X}_j^1 \setminus \mathcal{X}_{jk}^1] \cap \mathcal{X}^0)&+\nonumber\\
\frac{1}{s} \cdot (1-\gamma) \cdot  P(\mathcal{X}^0 \setminus \mathcal{X}_j^1)&-\nonumber\\
\frac{s-1}{s} \cdot  \gamma \cdot  P(\mathcal{X}_{jk}^1 \cap \mathcal{X}^1)&-\nonumber\\
\frac{s-1}{s} \cdot (1-\gamma) \cdot  P(\mathcal{X}_{jk}^1 \cap \mathcal{X}^0)&-\nonumber\\
(1 - \gamma) \cdot  P([\mathcal{X}_j^1 \setminus \mathcal{X}_{jk}^1] \cap \mathcal{X}^1)&-\nonumber\\
\gamma \cdot P([\mathcal{X}_j^1 \setminus \mathcal{X}_{jk}^1] \cap \mathcal{X}^0).
\end{eqnarray}
Since $(\mathcal{X}_j^1 \setminus \mathcal{X}_{jk}^1) \cap (\mathcal{X}^1 \setminus \mathcal{X}_j^1) = \emptyset$ and $(\mathcal{X}_j^1 \setminus \mathcal{X}_{jk}^1) \cup (\mathcal{X}^1 \setminus \mathcal{X}_j^1) = \mathcal{X}^1 \setminus \mathcal{X}_{jk}^1$, we can simplify to:
\begin{eqnarray}
\frac{1}{s} \cdot \gamma \cdot P(\mathcal{X}^1 \setminus \mathcal{X}_{jk}^1)&+\nonumber\\
\frac{1}{s} \cdot (1-\gamma) \cdot  P(\mathcal{X}^0 \setminus \mathcal{X}_{jk}^1)&-\nonumber\\
\frac{s-1}{s} \cdot  \gamma \cdot  P(\mathcal{X}_{jk}^1 \cap \mathcal{X}^1)&-\nonumber\\
\frac{s-1}{s} \cdot (1-\gamma) \cdot  P(\mathcal{X}_{jk}^1 \cap \mathcal{X}^0)&-\nonumber\\
(1 - \gamma) \cdot  P([\mathcal{X}_j^1 \setminus \mathcal{X}_{jk}^1] \cap \mathcal{X}^1)&-\nonumber\\
\gamma \cdot P([\mathcal{X}_j^1 \setminus \mathcal{X}_{jk}^1] \cap \mathcal{X}^0).
\end{eqnarray}
\end{proof}

\begin{mylemma}\label{lemma:expected_reward_include}
Consider a literal $l_k$ and clause $C_j^\omega$. The expected payoff of including $l_k$ in $C_j^\omega$ in the Tsetlin Machine game (Definition \ref{def:tsetlin_machine_game}) is: \begin{eqnarray}
\frac{s-1}{s} \cdot  \gamma \cdot  P(\mathcal{X}_{jk}^1 \cap \mathcal{X}^1) +
\frac{s-1}{s} \cdot (1-\gamma) \cdot  P(\mathcal{X}_{jk}^1 \cap \mathcal{X}^0)\nonumber&-\\
\frac{1}{s} \cdot \gamma \cdot P(\mathcal{X}^1 \setminus \mathcal{X}_{jk}^1) -
\frac{1}{s} \cdot (1-\gamma) \cdot  P(\mathcal{X}^0 \setminus \mathcal{X}_{jk}^1).&\label{eqn:expected_reward_include}
\end{eqnarray}
\end{mylemma}

\begin{proof}
The expected payoff of the 
\emph{Include} action can be derived from Table \ref{table:expected_feedback_include}:
\begin{eqnarray}
\frac{s-1}{s} \cdot  \gamma \cdot  P(\mathcal{X}_{jk}^1 \cap \mathcal{X}^1)&+\nonumber\\
\frac{s-1}{s} \cdot (1-\gamma) \cdot  P(\mathcal{X}_{jk}^1 \cap \mathcal{X}^0)&-\nonumber\\
\frac{1}{s} \cdot \gamma \cdot P([\mathcal{X}_j^1 \setminus \mathcal{X}_{jk}^1] \cap \mathcal{X}^1)&-\nonumber\\
\frac{1}{s} \cdot \gamma \cdot P(\mathcal{X}^1 \setminus \mathcal{X}_j^1)&-\nonumber\\
\frac{1}{s} \cdot (1-\gamma) \cdot P([\mathcal{X}_j^1 \setminus \mathcal{X}_{jk}^1] \cap \mathcal{X}^0)&-\nonumber\\
\frac{1}{s} \cdot (1-\gamma) \cdot  P(\mathcal{X}^0 \setminus \mathcal{X}_j^1),
\end{eqnarray}
which again simplifies to:
\begin{eqnarray}
\frac{s-1}{s} \cdot  \gamma \cdot  P(\mathcal{X}_{jk}^1 \cap \mathcal{X}^1)&+\nonumber\\
\frac{s-1}{s} \cdot (1-\gamma) \cdot  P(\mathcal{X}_{jk}^1 \cap \mathcal{X}^0)&-\nonumber\\
\frac{1}{s} \cdot \gamma \cdot P(\mathcal{X}^1 \setminus \mathcal{X}_{jk}^1)&-\nonumber\\
\frac{1}{s} \cdot (1-\gamma) \cdot  P(\mathcal{X}^0 \setminus \mathcal{X}_{jk}^1).
\end{eqnarray}
\end{proof}

\subsection{Tsetlin Automaton Convergence Criteria}

Most bandit algorithms attempt to learn the action with the largest expected payoff. A Tsetlin Automaton, however, can only learn this action if the expected payoff of the action is positive (i.e., the probability of receiving a reward is larger than the probability of receiving a penalty)~\cite{Narendra1989}. If the latter criterion is not fulfilled, the Tsetlin Automaton instead moves towards its center states ($\phi_N$ and $\phi_{N+1}$), that is, towards indecision. This property is crucial for eliminating sub-optimal Nash equilibria in Tsetlin Machine learning, and is one of the reasons why it is not straightforward to replace the Tsetlin Automaton with another bandit algorithm, without loss in learning accuracy. So, in order to have a stable Nash equilibrium in Tsetlin Machine learning, each Tsetlin Automaton action in the equilibrium must not only have the largest expected payoff, the expected payoff must also be positive. 
\begin{mylemma}\label{lemma:ta_convergence}
By increasing the number of states, a two-action Tsetlin Automaton converges to performing an action $\alpha_z, z \in \{0,1\},$ with probability arbitrarily close to unity iff:
\begin{enumerate}
\item The action has positive expected payoff: $E[R|\alpha_z] > 0$.
\item The action has the largest expected payoff: 
$E[R|\alpha_z] > E[R|\alpha_{1-z}]$.
\end{enumerate}
\end{mylemma}
\begin{proof}
The lemma follows trivially from the Tsetlin Automaton convergence proof found in~\cite{Narendra1989}.
\end{proof}

\subsection{Nash Equilibria Without Noise}

\begin{mylemma}\label{lemma:nash_equilibrium_without_noise}
Consider a pattern recognition problem per Definition \ref{def:problem_definition} without noise, $P(y \ne \omega | X^\omega) = 1 - \gamma = 0$,  and a Tsetlin Machine game per Definition \ref{def:tsetlin_machine_game} with a single clause $C^1(X)$ of positive polarity. The sub-patterns $Q_h^1(X)$, $h \in \{1,\ldots n/2\},$ of the pattern recognition problem are then all Nash equilibria of the game. Additionally, the expected payoff of each player action in each equilibrium is positive, hence not rejected by the Tsetlin Automata players.
\end{mylemma}

\begin{proof}
We first analyse the case when the \emph{Exclude} action is optimal, before addressing the \emph{Include} action.

\textbf{Exclude is Optimal.} From Property 1 and Property 2 of  Definition \ref{def:problem_definition}, we have $P(\mathcal{X}_{jk}^1 | \mathcal{X}^1) < \frac{1}{s} < P(\mathcal{X}_j^1 | \mathcal{X}^1) \Leftrightarrow \frac{P(\mathcal{X}_{jk}^1 \cap \mathcal{X}^1)}{P(\mathcal{X}^1)} < \frac{1}{s} < \frac{P(\mathcal{X}_j^1 \cap \mathcal{X}^1)}{P(\mathcal{X}^1)}$, which simplifies to $\frac{P(\mathcal{X}_{jk}^1)}{P(\mathcal{X}^1)} < \frac{1}{s} < \frac{P(\mathcal{X}_j^1)}{P(\mathcal{X}^1)}$ because of Property 4. From Property 4 we know in particular that $[\mathcal{X}_j^1 \setminus \mathcal{X}_{jk}^1] \cap \mathcal{X}^0 = \emptyset$. Under these conditions, we must verify that \emph{Exclude} is also the preferred action according to Lemma \ref{lemma:ta_convergence}.

We first verify that the expected payoff of \emph{Exclude} (Eqn. \ref{eqn:expected_reward_exclude}) is positive to accommodate for Criterion 1 of Lemma \ref{lemma:ta_convergence}. We here assume no noise, i.e., $\gamma = 1$, and thus can remove the addends containing the factor $(1-\gamma)$ from Eqn. \ref{eqn:expected_reward_exclude}. We further know that $P(\mathcal{X}_j^1 \setminus \mathcal{X}_{jk}^1] \cap \mathcal{X}^0) = 0$ because $[\mathcal{X}_j^1 \setminus \mathcal{X}_{jk}^1] \cap \mathcal{X}^0 = \emptyset$. Hence, we can remove addends including that factor as well, ending up with:
\begin{eqnarray}
\frac{1}{s} \cdot P(\mathcal{X}^1 \setminus \mathcal{X}_{jk}^1) -
\frac{s-1}{s} \cdot  P(\mathcal{X}_{jk}^1 \cap \mathcal{X}^1) & > & 0.
\end{eqnarray}
Because $\mathcal{X}_j^1$ is a subset of $\mathcal{X}^1$ (Property 4), we have:
\begin{eqnarray}
\frac{1}{s} \cdot P(\mathcal{X}^1 \setminus \mathcal{X}_{jk}^1) -
\frac{s-1}{s} \cdot  P(\mathcal{X}_{jk}^1) & > & 0.
\end{eqnarray}
Dividing by $P(\mathcal{X}^1)$ on each side we get:
\begin{eqnarray}
\frac{1}{s} \cdot \frac{P(\mathcal{X}^1 \setminus \mathcal{X}_{jk}^1)}{P(\mathcal{X}^1)} -
\frac{s-1}{s} \cdot \frac{ P(\mathcal{X}_{jk}^1)}{P(\mathcal{X}^1)} & > & 0.
\end{eqnarray}
With $\frac{P(\mathcal{X}_{jk}^1)}{P(\mathcal{X}^1)} < \frac{1}{s}$, we have $\frac{P(\mathcal{X}^1 \setminus \mathcal{X}_{jk}^1)}{P(\mathcal{X}^1)} > \frac{s-1}{s}$. This clearly entails that the left side is positive.

We next verify that the expected payoff of \emph{Exclude} is larger than that of \emph{Include} to ensure Criterion 2 of Lemma \ref{lemma:ta_convergence}. Performing the same simplifications as above, however, for Eqn. \ref{eqn:expected_reward_include}, we end up with the following condition:
\begin{eqnarray}
\frac{s-1}{s} \cdot \frac{P(\mathcal{X}_{jk}^1)}{P(\mathcal{X}^1)} -
\frac{1}{s} \cdot \frac{P(\mathcal{X}^1 \setminus \mathcal{X}_{jk}^1)}{P(\mathcal{X}^1)} &<& 0.
\end{eqnarray}
For the same reasons that the expected payoff of the \emph{Exclude} action was positive, the expected payoff of \emph{Include} is negative, which clearly is inferior.

\textbf{Include is Optimal.} Conversely, given that \emph{Include} is optimal, we have: $\frac{P(\mathcal{X}_{jk}^1)}{P(\mathcal{X}^1)} > \frac{1}{s}$ (Property 1) and $\mathcal{X}_{jk}^1 \cap \mathcal{X}^0 = \emptyset$ (from Property 4). With no noise we can again simplify the calculation of the expected payoff of \emph{Include}, reducing Eqn. \ref{eqn:expected_reward_include} to:
\begin{eqnarray}
\frac{s-1}{s} \cdot \frac{P(\mathcal{X}_{jk}^1)}{P(\mathcal{X}^1)} -
\frac{1}{s} \cdot \frac{P(\mathcal{X}^1 \setminus \mathcal{X}_{jk}^1)}{P(\mathcal{X}^1)} &>& 0.
\end{eqnarray}
With $\frac{P(\mathcal{X}_{jk}^1)}{P(\mathcal{X}^1)} > \frac{1}{s}$ (Property 1), we must have $\frac{P(\mathcal{X}^1 \setminus \mathcal{X}_{jk}^1)}{P(\mathcal{X}^1)} < \frac{s-1}{s}$. Accordingly, the expected payoff of \emph{Include} is positive.

We now turn to investigating whether the expected payoff of \emph{Include} is larger than the expected payoff of \emph{Exclude} (Criterion 2 of Lemma \ref{lemma:ta_convergence}), which we show by verifying that the expected payoff of \emph{Exclude} is negative. We can again simplify Eqn. \ref{eqn:expected_reward_exclude}, however, this time without knowing that $[\mathcal{X}_j^1 \setminus \mathcal{X}_{jk}^1] \cap \mathcal{X}^0 = \emptyset$:
\begin{eqnarray}
\frac{1}{s} \cdot P(\mathcal{X}^1 \setminus \mathcal{X}_{jk}^1) -
\frac{s-1}{s} \cdot P(\mathcal{X}_{jk}^1) -
P([\mathcal{X}_j^1 \setminus \mathcal{X}_{jk}^1] \cap \mathcal{X}^0)
&<& 0.
\end{eqnarray}
Dividing by $P(\mathcal{X}^1)$ on each side we get:
\begin{eqnarray}
\frac{1}{s} \cdot \frac{P(\mathcal{X}^1 \setminus \mathcal{X}_{jk}^1)}{P(\mathcal{X}^1)} -
\frac{s-1}{s} \cdot \frac{P(\mathcal{X}_{jk}^1)}{P(\mathcal{X}^1)} -
\frac{P([\mathcal{X}_j^1 \setminus \mathcal{X}_{jk}^1] \cap \mathcal{X}^0)}{P(\mathcal{X}^1)}
&<& 0.
\end{eqnarray}
Again, with $\frac{P(\mathcal{X}_{jk}^1)}{P(\mathcal{X}^1)} > \frac{1}{s}$, we have $\frac{P(\mathcal{X}^1 \setminus \mathcal{X}_{jk}^1)}{P(\mathcal{X}^1)} < \frac{s-1}{s}$. From this result alone, the left side is negative. Additionally, the addend $\frac{P([\mathcal{X}_j^1 \setminus \mathcal{X}_{jk}^1] \cap \mathcal{X}^0)}{P(\mathcal{X}^1)}$ is negative as well, reducing the expected payoff of \emph{Exclude} even further. So, changing from \emph{Include} to \emph{Exclude} gives a net expected loss.
\end{proof}

\begin{mylemma}\label{lemma:nash_equilibrium_without_noise_ii}
Consider a pattern recognition problem per Definition \ref{def:problem_definition} without noise, $P(y \ne \omega | X^\omega) = 1 - \gamma = 0$, and a Tsetlin Machine game per Definition \ref{def:tsetlin_machine_game} with a single clause $C^1(X)$ of positive polarity.  If clause $C^1(X)$ deviates from all of the sub-patterns $Q_h^1(X)$, $h \in \{1,\ldots n/2\},$ of the pattern recognition problem, it is either not a Nash equilibrium in the game or it is produced by an action with negative expected payoff.
\end{mylemma}

\begin{proof}
There are three kinds of sub-optimal configurations for a clause: (i) The clause is too infrequent; (ii) The clause is too frequent; or (iii) The clause outputs $1$ for input vectors from $\mathcal{X}^0$.   We must show that none of these fulfill both criteria of Lemma \ref{lemma:ta_convergence}.

In case (i) there exists at least one literal $l_k$ that has been included in the clause, for which $\frac{P(\mathcal{X}_{jk}^1 \cap \mathcal{X}^1)}{P(\mathcal{X}^1)} < \frac{1}{s}$. That is, less than $\frac{1}{s}$ of $\mathcal{X}^1$  is covered by the clause, in frequency. It is sufficient to show that the expected payoff of \emph{Include} (Eqn. \ref{eqn:expected_reward_include}) is negative for $\gamma=1$:
\begin{eqnarray}
\frac{s-1}{s} \cdot P(\mathcal{X}_{jk}^1 \cap \mathcal{X}^1) -
\frac{1}{s} \cdot P(\mathcal{X}^1 \setminus \mathcal{X}_{jk}^1) &<& 0.
\end{eqnarray}
Dividing by $P(\mathcal{X}^1)$ on each side we get:
\begin{eqnarray}
\frac{s-1}{s} \cdot \frac{P(\mathcal{X}_{jk}^1 \cap \mathcal{X}^1)}{P(\mathcal{X}^1)} -
\frac{1}{s} \cdot \frac{P(\mathcal{X}^1 \setminus \mathcal{X}_{jk}^1)}{P(\mathcal{X}^1)} &<& 0.
\end{eqnarray}
With $\frac{P(\mathcal{X}_{jk}^1\cap \mathcal{X}^1)}{P(\mathcal{X}^1)} < \frac{1}{s}$, we must have $\frac{P(\mathcal{X}^1 \setminus \mathcal{X}_{jk}^1)}{P(\mathcal{X}^1)} > \frac{s-1}{s}$.  We can thus conclude that the expected payoff of \emph{Include} is negative.

We now turn to case (ii), that is, the case where the clause is too frequent. Then there must exist an excluded literal $l_k$ such that $\frac{P(\mathcal{X}_{jk}^1 \cap \mathcal{X}^1)}{P(\mathcal{X}^1)} > \frac{1}{s}$. It is then sufficient to show that the expected payoff of \emph{Exclude} is negative:
\begin{eqnarray}
\frac{1}{s} \cdot \frac{P(\mathcal{X}^1 \setminus \mathcal{X}_{jk}^1)}{P(\mathcal{X}^1)} -
\frac{s-1}{s} \cdot \frac{P(\mathcal{X}_{jk}^1 \cap \mathcal{X}^1)}{P(\mathcal{X}^1)} -
\frac{P([\mathcal{X}_j^1 \setminus \mathcal{X}_{jk}^1] \cap \mathcal{X}^0)}{P(\mathcal{X}^1)} &<&0.
\end{eqnarray}
With $\frac{P(\mathcal{X}_{jk}^1 \cap \mathcal{X}^1)}{P(\mathcal{X}^1)} > \frac{1}{s}$, we have $\frac{P(\mathcal{X}^1 \setminus \mathcal{X}_{jk}^1)}{P(\mathcal{X}^1)} < \frac{s-1}{s}$.  Again,  the left side is negative.

Apart from the sub-optimal configurations where too many or too few literals are included in the clause, we must also consider case (iii), that is, scenarios where $P([\mathcal{X}_j^1 \setminus \mathcal{X}_{jk}^1] \cap \mathcal{X}^0) > 0$. This means that the corresponding clause $C_j^1(X)$ outputs $1$ for input vectors $X\in [\mathcal{X}_j^1 \setminus \mathcal{X}_{jk}^1] \cap \mathcal{X}^0$. Accordingly, it outputs $1$ when $P(y = 1|X) < P(y=0|X)$, which is sub-optimal and violates Property 4 of Definition \ref{def:problem_definition}. It must then clearly exist a literal $l_k$ that has been excluded from the clause and that is zero for input vectors $X\in [\mathcal{X}_j^1 \setminus \mathcal{X}_{jk}^1] \cap \mathcal{X}^0$. If not, an optimal configuration would not exist, which it does by problem definition. Including $l_k$ in the clause would make the clause output $0$ for input vectors $X\in [\mathcal{X}_j^1 \setminus \mathcal{X}_{jk}^1] \cap \mathcal{X}^0$. This will thus eliminate $X\in [\mathcal{X}_j^1 \setminus \mathcal{X}_{jk}^1] \cap \mathcal{X}^0$ from the input space captured by the clause. Accordingly, we must ensure that excluding $l_k=0$ gives a negative expected payoff:
\begin{eqnarray}
\frac{1}{s} \cdot P(\mathcal{X}^1 \setminus \mathcal{X}_{jk}^1) -
\frac{s-1}{s} \cdot  P(\mathcal{X}_{jk}^1 \cap \mathcal{X}^1) -
P([\mathcal{X}_j^1 \setminus \mathcal{X}_{jk}^1] \cap \mathcal{X}^0) &<& 0.
\end{eqnarray}
We can assume that $\frac{1}{s} \cdot P(\mathcal{X}^1 \setminus \mathcal{X}_{jk}^1)$ and $\frac{s-1}{s} \cdot P(\mathcal{X}_{jk}^1 \cap \mathcal{X}^1)$ are roughly equal. If they are not, we have already determined in the previous two paragraphs that the resulting clause is not a Nash equilibrium. Thus, what remains is to show that $P([\mathcal{X}_j^1 \setminus \mathcal{X}_{jk}^1] \cap \mathcal{X}^0)>0$, which is stated by case (iii) itself.
\end{proof}

\begin{mytheorem}\label{theorem:optimality_without_noise}
Consider a pattern recognition problem per Definition \ref{def:problem_definition} without noise, $P(y \ne \omega | X^\omega) = 1 - \gamma = 0$, and a Tsetlin Machine game per Definition \ref{def:tsetlin_machine_game} with a single clause $C^1(X)$ of positive polarity. The sub-patterns $Q_h^1(X)$, $h \in \{1,\ldots n/2\},$ are then the only Nash equilibria of the game, where the expected payoff of each player action is positive, hence not rejected by the Tsetlin Automata players.
\end{mytheorem}
\begin{proof}
This theorem follows from Lemma \ref{lemma:ta_convergence}, Lemma \ref{lemma:nash_equilibrium_without_noise} and Lemma \ref{lemma:nash_equilibrium_without_noise_ii}.
\end{proof}

\subsection{Nash Equilibria with Static Noise}

\begin{mylemma}\label{lemma:nash_equilibrium_with_static_noise}
Consider a pattern recognition problem per Definition \ref{def:problem_definition} with static noise, $P(y \ne \omega | X^\omega) = 1 - \gamma, 0.5 < \gamma < 1.0$, and output classes of equal size, $P(X^0)=P(X^1)=0.5$. Consider further a Tsetlin Machine game per Definition \ref{def:tsetlin_machine_game} with a single clause $C^1(X)$ of positive polarity. Each sub-pattern $Q_h^1(X)$, $h \in \{1,\ldots n/2\},$ of the pattern recognition problem is a Nash equilibrium in the game when $s$ is scaled by a factor $t$,
\begin{eqnarray}
 \frac{b_1}{\gamma} &< t < & \frac{b_2}{\frac{b_2}{b_3} (1 - \gamma) + 2 ( \gamma - 0.5)},
\end{eqnarray}
where, $b_2 = \mathrm{min}_{j,k} \frac{P(\mathcal{X}^1)}{P(\mathcal{X}_{jk}^1)\cdot s}, b_2 > 1,$ and $b_3 = \mathrm{max}_{j,k} \frac{P(\mathcal{X}^1)}{P(\mathcal{X}_j^1) \cdot s}, 0 < b_3 < 1,$ define a worst-case upper bound for $t$ when \emph{Exclude} is optimal. Further, $b_1 = \mathrm{max}_{j,k} \frac{P(\mathcal{X}^1)}{P(\mathcal{X}_{jk}^1) \cdot s}, 0 < b_1 < 1,$ defines a worst-case lower-bound when \emph{Include} is optimal. Additionally, the expected payoff of each player action in each equilibrium is positive, hence not rejected by the Tsetlin Automata players.
\end{mylemma}
As seen, increasing noise (reduced $\gamma$) tightens the lower bound, but loosens the upper. Further, an increasing $b_2$ loosens the upper bound, while a decreasing $b_1$ loosens the lower bound.

\begin{proof}
We first analyse the case when \emph{Exclude} is optimal, before addressing \emph{Include}.

\textbf{Exclude is Optimal.}
Since \emph{Exclude} is optimal, we have (follows from Definition \ref{def:problem_definition}): $P(\mathcal{X}_{jk}^1 \cap \mathcal{X}^0)=0$, $P([\mathcal{X}_j^1 \setminus \mathcal{X}_{jk}^1] \cap \mathcal{X}^0)=0$,  $P(\mathcal{X}_{jk}^1 \cap \mathcal{X}^1)=P(\mathcal{X}_{jk}^1)$, $P([\mathcal{X}_j^1 \setminus \mathcal{X}_{jk}^1] \cap \mathcal{X}^1) = P(\mathcal{X}_j^1 \setminus \mathcal{X}_{jk}^1)$, and $P(\mathcal{X}^0 \setminus \mathcal{X}_{jk}^1)=P(\mathcal{X}^0)$. We consider Criterion 1 of Lemma \ref{lemma:ta_convergence} first, i.e., that the expected payoff from Eqn.~\ref{eqn:expected_reward_exclude} is positive, which simplifies to:
\begin{eqnarray}
\frac{1}{s} \cdot \gamma \cdot P(\mathcal{X}^1 \setminus \mathcal{X}_{jk}^1) +
\frac{1}{s} \cdot (1-\gamma) \cdot P(\mathcal{X}^0) &-&\nonumber\\
\frac{s-1}{s} \cdot  \gamma \cdot  P(\mathcal{X}_{jk}^1) -
(1 - \gamma) \cdot  P(\mathcal{X}_j^1 \setminus \mathcal{X}_{jk}^1)&>&0.
\end{eqnarray}
Dividing by $P(\mathcal{X}^1)$ on each side, we get:
\begin{eqnarray}
\frac{1}{s} \cdot \gamma \cdot \frac{P(\mathcal{X}^1 \setminus \mathcal{X}_{jk}^1)}{P(\mathcal{X}^1)} +
\frac{1}{s} \cdot (1-\gamma) \cdot  \frac{P(\mathcal{X}^0)}{P(\mathcal{X}^1)} &-&\nonumber\\
\frac{s - 1}{s} \cdot  \gamma \cdot  \frac{P(\mathcal{X}_{jk}^1)}{P(\mathcal{X}^1)} -
(1 - \gamma) \cdot  \frac{P(\mathcal{X}_j^1 \setminus \mathcal{X}_{jk}^1)}{P(\mathcal{X}^1)} &>&0.
\end{eqnarray}
Clearly, to maintain positive expected payoff after the introduction of static noise, we need to compensate for the addends introduced by the noise, i.e., the $(1-\gamma$)-addends. To this end, we artificially scale up the specificity parameter $s$ by a factor $t$. Simplifying further, we replace $\frac{P(\mathcal{X}^0)}{P(\mathcal{X}^1)}$ with $1$:
\begin{eqnarray}
\frac{1}{ts} \cdot \gamma \cdot \frac{P(\mathcal{X}^1 \setminus \mathcal{X}_{jk}^1)}{P(\mathcal{X}^1)} +
\frac{1}{ts} \cdot (1-\gamma)  &-&\nonumber\\
\frac{ts - 1}{ts} \cdot  \gamma \cdot  \frac{P(\mathcal{X}_{jk}^1)}{P(\mathcal{X}^1)} -
(1 - \gamma) \cdot  \frac{P(\mathcal{X}_j^1 \setminus \mathcal{X}_{jk}^1)}{P(\mathcal{X}^1)} &>&0.\label{eqn:exclude_nash_noise}
\end{eqnarray}
Since we consider the case where excluding literal $l_k$ is optimal, we have $b_2 > 1$ such that: $\frac{P(\mathcal{X}_{jk}^1)}{P(\mathcal{X}^1)} = \frac{1}{b_2 s} \Rightarrow b_2 = \frac{P(\mathcal{X}^1)}{P(\mathcal{X}_{jk}^1)\cdot s}$. That is, $b_2$ is an unknown factor that measures exactly how much including $l_k$ constrains the clause beyond $\frac{1}{s}$, according to Property 1 and Property 2 of Definition \ref{def:problem_definition}. This in turn, means that $\frac{P(\mathcal{X}^1 \setminus \mathcal{X}_{jk}^1)}{P(\mathcal{X}^1)} = 1 - \frac{P(\mathcal{X}_{jk}^1)}{P(\mathcal{X}^1)} = \frac{b_2 s - 1}{b_2 s}$. We finally introduce $b_3, 0 < b_3 < 1$. The role of $b_3$ is to exactly determine how much excluding literal $l_k$ loosens up the clause beyond $\frac{1}{s}$: $\frac{P(\mathcal{X}_j^1)}{P(\mathcal{X}^1)} = \frac{1}{b_3 s} \Rightarrow b_3 = \frac{P(\mathcal{X}^1)}{P(\mathcal{X}_j^1) \cdot s}$. This leads to $\frac{P(\mathcal{X}_j^1 \setminus \mathcal{X}_{jk}^1)}{P(\mathcal{X}^1)} = \frac{P(\mathcal{X}_j^1)}{P(\mathcal{X}^1)} - \frac{P(\mathcal{X}_{jk}^1)}{P(\mathcal{X}^1)} = \frac{1}{b_3 s} - \frac{1}{b_2 s}$. With these amendments, we revisit the condition that the expected payoff of \emph{Exclude} must be positive:
\begin{eqnarray}
\frac{1}{t s} \cdot \gamma \cdot \frac{b_2 s - 1}{b_2 s} +
\frac{1}{t s} \cdot (1-\gamma)  -
\frac{t s - 1}{t s} \cdot \gamma \cdot  \frac{1}{b_2 s} -
(1 - \gamma) \cdot \left(\frac{1}{b_3 s} - \frac{1}{b_2 s}\right) &>&0.
\end{eqnarray}
Through a sequence of standard algebraic operations, the above condition can be simplified as follows:
\begin{eqnarray}
\gamma \cdot \frac{b_2 s - 1}{b_2 s} +
(1-\gamma)  -
(t s - 1) \cdot \gamma \cdot  \frac{1}{b_2 s} -
ts \cdot (1 - \gamma) \cdot \left(\frac{1}{b_3 s} - \frac{1}{b_2 s}\right) &>&0\\
%
%
\gamma \cdot \frac{b_2 s - 1}{b_2 s} +
(1-\gamma)  -
(t s - 1) \cdot \gamma \cdot  \frac{1}{b_2 s} -
ts \cdot (1 - \gamma) \cdot \frac{b_2 - b_3}{b_2 b_3 s} &>&0\\
%
%
\gamma \cdot (b_2 s - 1) +
b_2 s \cdot (1-\gamma)  -
(t s - 1) \cdot \gamma -
ts \cdot (1 - \gamma) \cdot \frac{b_2-b_3}{b_3} &>&0\\
%
b_2 s \gamma  - \gamma +
b_2 s - b_2 s \gamma  -
t s \gamma + \gamma -
ts \cdot (1 - \gamma) \cdot \frac{b_2-b_3}{b_3}
&>&0\\
%
b_2 \gamma  +
b_2 - b_2 \gamma  -
t \gamma -
t \cdot (1 - \gamma) \cdot \frac{b_2-b_3}{b_3}
&>&0\\
%
b_2 \gamma  +
b_2 - b_2 \gamma  -
t \gamma -
t \cdot (1 - \gamma) \cdot \left(\frac{b_2}{b_3}-1\right)
&>&0\\
%
b_2 \gamma  +
b_2 - b_2 \gamma  -
t \gamma -
\left(t \cdot (1 - \gamma) \cdot \frac{b_2}{b_3}- t \cdot (1 - \gamma) \right)
&>&0\\
%
b_2 \gamma  +
b_2 - b_2 \gamma  -
t \gamma -
t \cdot (1 - \gamma) \cdot \frac{b_2}{b_3} + t \cdot (1 - \gamma)
&>&0\\
%
b_2 -
t \gamma -
t \cdot (1 - \gamma) \cdot \frac{b_2}{b_3} + t \cdot (1 - \gamma)
&>&0\\
%
b_2 -
t \gamma -
\frac{b_2}{b_3}t + \frac{b_2}{b_3}t \gamma + t - t \gamma
&>&0\\
%
b_2 -
t \left(\gamma +
\frac{b_2}{b_3} - \frac{b_2}{b_3}\gamma - 1 + \gamma\right)
&>&0\\
%
\frac{b_2}
{\gamma +
\frac{b_2}{b_3} - \frac{b_2}{b_3}\gamma - 1 + \gamma}
&>&t\\
%
\frac{b_2}
{
\frac{b_2}{b_3} (1 - \gamma) + 2 ( \gamma - \frac{1}{2})}
&>&t
\end{eqnarray}
%
This means that it is possible to increase $t$ up to a certain boundary, weakening reinforcement of \emph{Exclude}, without making the expected payoff of \emph{Exclude} turn negative.

We now show that the expected payoff of \emph{Include} is negative, fulfilling Criterion 2 of Lemma~\ref{lemma:ta_convergence}:
\begin{eqnarray}
\frac{s-1}{s} \cdot  \gamma \cdot  P(\mathcal{X}_{jk}^1 \cap \mathcal{X}^1) +
\frac{s-1}{s} \cdot (1-\gamma) \cdot  P(\mathcal{X}_{jk}^1 \cap \mathcal{X}^0)&-\nonumber\\
\frac{1}{s} \cdot \gamma \cdot P(\mathcal{X}^1 \setminus \mathcal{X}_{jk}^1) -
\frac{1}{s} \cdot (1-\gamma) \cdot  P(\mathcal{X}^0 \setminus \mathcal{X}_{jk}^1)&<&0.
\end{eqnarray}
Since \emph{Exclude} is optimal, we can simplify as follows (per Definition \ref{def:problem_definition}):
\begin{eqnarray}
\frac{s-1}{s} \cdot  \gamma \cdot  P(\mathcal{X}_{jk}^1)-
\frac{1}{s} \cdot \gamma \cdot P(\mathcal{X}^1 \setminus \mathcal{X}_{jk}^1)-
\frac{1}{s} \cdot (1-\gamma) \cdot  P(\mathcal{X}^0)&<&0.
\end{eqnarray}
Introducing the scaling factor $t > 1$ for $s$ and dividing by $P(\mathcal{X}^1)$ we get:
\begin{eqnarray}
\frac{ts-1}{ts} \cdot  \gamma \cdot  \frac{P(\mathcal{X}_{jk}^1)}{P(\mathcal{X}^1)} -
\frac{1}{ts} \cdot \gamma \cdot \frac{P(\mathcal{X}^1 \setminus \mathcal{X}_{jk}^1)}{P(\mathcal{X}^1)}-
\frac{1}{ts} \cdot (1-\gamma) \cdot  \frac{P(\mathcal{X}^0)}{P(\mathcal{X}^1)}&<&0.
\end{eqnarray}
Because we have established that Eqn. \ref{eqn:exclude_nash_noise} is positive, it follows that the above equation is negative. This is due to the fact Eqn. \ref{eqn:exclude_nash_noise} is still positive after removing the last addend, and the above equation is simply the negated version of the latter.

\textbf{Include is Optimal.}
To fulfill Criterion 1 of Lemma \ref{lemma:ta_convergence} we must show that the expected payoff of \emph{Include} is positive:
\begin{eqnarray}
\frac{s-1}{s} \cdot  \gamma \cdot  P(\mathcal{X}_{jk}^1 \cap \mathcal{X}^1) +
\frac{s-1}{s} \cdot (1-\gamma) \cdot  P(\mathcal{X}_{jk}^1 \cap \mathcal{X}^0)&-\nonumber\\
\frac{1}{s} \cdot \gamma \cdot P(\mathcal{X}^1 \setminus \mathcal{X}_{jk}^1) -
\frac{1}{s} \cdot (1-\gamma) \cdot  P(\mathcal{X}^0 \setminus \mathcal{X}_{jk}^1)&>&0.
\end{eqnarray}
Since \emph{Include} is optimal, we have: $P(\mathcal{X}_{jk}^1 \cap \mathcal{X}^0) = 0$, $P(\mathcal{X}_{jk}^1 \cap \mathcal{X}^1) = P(\mathcal{X}_{jk}^1)$  and $P(\mathcal{X}^0 \setminus \mathcal{X}_{jk}^1)=P(\mathcal{X}^0)$, again per Property 4 of Definition \ref{def:problem_definition}. Accordingly, we can simplify the above expression to:
\begin{eqnarray}
\frac{s-1}{s} \cdot  \gamma \cdot  P(\mathcal{X}_{jk}^1)-
\frac{1}{s} \cdot \gamma \cdot P(\mathcal{X}^1 \setminus \mathcal{X}_{jk}^1)-
\frac{1}{s} \cdot (1-\gamma) \cdot  P(\mathcal{X}^0)&>&0.
\end{eqnarray}
To compensate for the addends introduced by noise, we again increase the $s$-parameter by artificially multiplying it with a factor $t > 1$. Dividing by $P(\mathcal{X}^1)$ we get:
\begin{eqnarray}
\frac{ts-1}{ts} \cdot  \gamma \cdot  \frac{P(\mathcal{X}_{jk}^1)}{P(\mathcal{X}^1)} -
\frac{1}{ts} \cdot \gamma \cdot \frac{P(\mathcal{X}^1 \setminus \mathcal{X}_{jk}^1)}{P(\mathcal{X}^1)}-
\frac{1}{ts} \cdot (1-\gamma) \cdot  \frac{P(\mathcal{X}^0)}{P(\mathcal{X}^1)}&>&0.\label{eqn:include_nash_noise}
\end{eqnarray}
Now we introduce $0 < b_1 < 1$, which tells exactly how much larger $\frac{P(\mathcal{X}_{jk}^1)}{P(\mathcal{X}^1)}$ is compared to $\frac{1}{s}$, i.e., $\frac{P(\mathcal{X}_{jk}^1)}{P(\mathcal{X}^1)} = \frac{1}{b_1 s} \Rightarrow b_1 =\frac{P(\mathcal{X}^1)}{P(\mathcal{X}_{jk}^1) \cdot s}$  as determined by Property 1 and Property 2 of Definition \ref{def:problem_definition}, we can simplify the condition as follows, also replacing $\frac{P(\mathcal{X}^0)}{P(\mathcal{X}^1)}$ with $1$:
\begin{eqnarray}
\frac{ts-1}{ts} \cdot  \gamma \cdot \frac{1}{b_1s}-
\frac{1}{ts} \cdot \gamma \cdot \frac{b_1s-1}{b_1s}-
\frac{1}{ts} \cdot (1-\gamma)&>&0\\
\frac{ts-1}{t} \cdot  \gamma \cdot \frac{1}{b_1}-
\frac{1}{t} \cdot \gamma \cdot \frac{b_1s-1}{b_1}-
\frac{s}{t} \cdot (1-\gamma)&>&0\\
(ts-1) \cdot  \gamma -
\gamma \cdot (b_1s-1) -
b_1 s \cdot (1-\gamma)&>&0\\
ts\gamma - \gamma -
b_1s\gamma + \gamma -
b_1 s + b_1s \gamma&>&0\\
ts\gamma  -
b_1 s &>&0\\
\frac{b_1}{\gamma} &<&t
\end{eqnarray}
This leads to the following range for $t$, recovering a Nash equilibrium corresponding to the sub-patterns of the pattern recognition problem:
\begin{eqnarray}
\frac{b_1}{\gamma}
< &t& <
\frac{b_2}{\frac{b_2}{b_3} (1 - \gamma) + 2 ( \gamma - \frac{1}{2})}
\end{eqnarray}
That is, scaling $s$ by a factor $t$ allows us to recover from some degree of static noise, as governed by $b_1, b_2,$ and $b_3$.

We finally need to verify that the expected payoff of \emph{Exclude} is negative, in order to fulfill Criterion 2 of Lemma~\ref{lemma:ta_convergence}:
\begin{eqnarray}
\frac{1}{s} \cdot \gamma \cdot P(\mathcal{X}^1 \setminus \mathcal{X}_{jk}^1) +
\frac{1}{s} \cdot (1-\gamma) \cdot  P(\mathcal{X}^0) -
\frac{s-1}{s} \cdot  \gamma \cdot  P(\mathcal{X}_{jk}^1)&-\nonumber\\
(1 - \gamma) \cdot  P([\mathcal{X}_j^1 \setminus \mathcal{X}_{jk}^1] \cap \mathcal{X}^1) -
\gamma \cdot P([\mathcal{X}_j^1 \setminus \mathcal{X}_{jk}^1] \cap \mathcal{X}^0)& < & 0.
\end{eqnarray}
Again, introducing the scaling factor $t > 1$ for $s$ and dividing by $P(\mathcal{X}^1)$ we get:
\begin{eqnarray}
\frac{1}{ts} \cdot \gamma \cdot \frac{P(\mathcal{X}^1 \setminus \mathcal{X}_{jk}^1)}{P(\mathcal{X}^1)} +
\frac{1}{ts} \cdot (1-\gamma) \cdot  \frac{P(\mathcal{X}^0)}{P(\mathcal{X}^1)} -
\frac{ts-1}{ts} \cdot  \gamma \cdot  \frac{P(\mathcal{X}_{jk}^1)}{P(\mathcal{X}^1)}&-\nonumber\\
(1 - \gamma) \cdot \frac{P([\mathcal{X}_j^1 \setminus \mathcal{X}_{jk}^1] \cap \mathcal{X}^1)}{P(\mathcal{X}^1)} -
\gamma \cdot \frac{P([\mathcal{X}_j^1 \setminus \mathcal{X}_{jk}^1] \cap \mathcal{X}^0)}{P(\mathcal{X}^1)}& < & 0.
\end{eqnarray}
Since we have established that Eqn. \ref{eqn:include_nash_noise} is positive, it follows that the above equation is negative. This is due to the above equation simply being the negated version of the latter, subtracting two extra addends, clearly making it negative.
\end{proof}

\begin{mylemma}\label{lemma:nash_equilibrium_with_static_noise_ii}
Consider a pattern recognition problem per Definition \ref{def:problem_definition} with static noise, $P(y \ne \omega | X^\omega) = 1 - \gamma, 0.5 < \gamma < 1.0$, and output classes of equal size, $P(y=0)=P(y=1)=0.5$. Consider further a Tsetlin Machine game per Definition \ref{def:tsetlin_machine_game} with only a single clause $C^1(X)$ of positive polarity. If clause $C^1(X)$ deviates from all of the sub-patterns $Q_h^1(X)$, $h \in \{1,\ldots n/2\},$ of the pattern recognition problem, it is either not a Nash equilibrium in the game or it is produced by an action with negative payoff.
\end{mylemma}

\begin{proof}
For too infrequent or too frequent clauses, the proof follows the structure of the corresponding proof for the noise-free cases (i) and (ii). However, we need to introduce the scaling factor $t$, as demonstrated above.

Case (iii) requires some more scrutiny. That is, we need to show that when $P([\mathcal{X}_j^1 \setminus \mathcal{X}_{jk}^1] \cap \mathcal{X}^0) > 0$, the expected payoff of \emph{Exclude} is negative:
\begin{eqnarray}
\frac{1}{ts} \cdot \gamma \cdot P(\mathcal{X}^1 \setminus \mathcal{X}_{jk}^1)&+\nonumber\\
\frac{1}{ts} \cdot (1-\gamma) \cdot  P(\mathcal{X}^0)&-\nonumber\\
\frac{ts-1}{ts} \cdot  \gamma \cdot  P(\mathcal{X}_{jk}^1)&-\nonumber\\
(1 - \gamma) \cdot  P([\mathcal{X}_j^1 \setminus \mathcal{X}_{jk}^1] \cap \mathcal{X}^1)&-\nonumber\\
\gamma \cdot P([\mathcal{X}_j^1 \setminus \mathcal{X}_{jk}^1] \cap \mathcal{X}^0)&<&0.
\end{eqnarray}
Again we can assume that $\frac{ts-1}{ts} \cdot  \gamma \cdot  P(\mathcal{X}_{jk}^1) -
\frac{1}{ts} \cdot \gamma \cdot P(\mathcal{X}^1 \setminus \mathcal{X}_{jk}^1) -
\frac{1}{ts} \cdot (1-\gamma) \cdot  P(\mathcal{X}^0)$ is close to zero. Otherwise, we did not have a Nash equilibrium in the first place, due to case (i) and case (ii). Thus, we only need to verify that:
\begin{eqnarray}
(1 - \gamma) \cdot  P([\mathcal{X}_j^1 \setminus \mathcal{X}_{jk}^1] \cap \mathcal{X}^1) + \nonumber \gamma \cdot P([\mathcal{X}_j^1 \setminus \mathcal{X}_{jk}^1] \cap \mathcal{X}^0)&>&0.
\end{eqnarray}
The above condition is clearly true because we have $P([\mathcal{X}_j^1 \setminus \mathcal{X}_{jk}^1] \cap \mathcal{X}^0) > 0$ by case (iii) itself. Additionally, because we have noise, $(1 - \gamma) \cdot  P([\mathcal{X}_j^1 \setminus \mathcal{X}_{jk}^1] \cap \mathcal{X}^1)$ is strictly positive as well.
\end{proof}

\begin{mytheorem}\label{theorem:optimality_with_static_noise}
Consider a pattern recognition problem per Definition \ref{def:problem_definition} with static noise, $P(y \ne \omega | X^\omega) = 1 - \gamma, 0.5 < \gamma < 1.0$, and output classes of equal size, $P(y=0)=P(y=1)=0.5$. Consider further a Tsetlin Machine game per Definition \ref{def:tsetlin_machine_game} with only a single clause $C^1(X)$ of positive polarity. The sub-patterns $Q_h^1(X)$, $h \in \{1,\ldots n/2\},$ of the pattern recognition problem are then the only Nash equilibria of the game, where the expected payoff of each player action is positive, hence not rejected by the Tsetlin Automata players.
\end{mytheorem}
\begin{proof}
This theorem follows from Lemma Lemma \ref{lemma:ta_convergence}, Lemma \ref{lemma:nash_equilibrium_with_static_noise}, and Lemma \ref{lemma:nash_equilibrium_with_static_noise_ii}.
\end{proof}

\subsection{Coordination of Multiple Tsetlin Automata Teams}

\begin{mytheorem}\label{theorem:optimality_without_noise_system}
Consider a pattern recognition problem per Definition \ref{def:problem_definition} without noise, $P(y \ne \omega | X^\omega) = 1 - \gamma = 0$, and a Tsetlin Machine game per Definition \ref{def:tsetlin_machine_game} with $T \cdot n$ clauses. Then the only Nash equilibrium of the Tsetlin Machine game is a configuration where $T$ Tsetlin Machine clauses $C^\omega_j, \omega \in \{0, 1\}, j \in \{1, \ldots, T \cdot n / 2\}$, duplicate each sub-pattern $Q^\omega_h, \omega \in \{0, 1\}, h \in \{1, \ldots, n/2\}$.
\end{mytheorem}

\begin{proof}
We already known from Theorem \ref{theorem:optimality_without_noise} that if a clause $C^\omega_j$ is not duplicating any of the sub-patterns $Q^\omega_h$, the Tsetlin Automata of the clause are not in a Nash Equilibrium. We therefore only consider the case where all clauses $C^\omega_j$ are duplicating some sub-pattern $Q^\omega_h$. Let us first consider positive polarity clauses ($\omega=1$). Assume that more than $T$ of these clauses duplicate a particular sub-pattern $Q_h^1$. Then, clearly, there are fewer than $T$ clauses duplicating some other sub-pattern $Q_{h'}^1, h'\ne h$. The Tsetlin Automata that take part in duplicating $Q_h^1$ will then only receive Inaction-feedback when facing sub-pattern $Q_h^1$, as seen from Eqn. \ref{eqn:activation_type_i}. However, when they observe sub-pattern $Q_{h'}^1$ they will receive Type I Feedback with probability larger than $0$. Type I Feedback pulls the Tsetlin Automata away from $Q_{h}^1$ and towards $Q_{h'}^1$. Hence, we do not have a Nash equilibrium. For the same reasons, we do not have a Nash equilibrium when more than $T$ negative polarity clauses capture a sub-pattern $Q^0_h$ due to Eqn. \ref{eqn:activation_type_ib}. When there are $T$ clauses per sub-pattern $Q^\omega_h$, on the other hand, the whole system is in equilibrium due to Theorem \ref{theorem:optimality_without_noise} and Eqns. \ref{eqn:activation_type_i}-\ref{eqn:activation_type_iib}.
\end{proof}
The proof for static noise follows the above structure, however, relies on Theorem \ref{theorem:optimality_with_static_noise} instead. The proof is left out here for the sake of brevity.

\paragraph{Remark 1.} Note that a Tsetlin Machine in practice also learns robustly when the total number of clauses available is less than $T \cdot n$. This is because of the gradually increasing probability of receiving feedback the farther away one is from $T$/$-T$, which leads to a load balancing effect among the clauses. This effect dynamically distributes the clauses among the available sub-patterns.

\paragraph{Remark 2.} A traditional Learning Automata approach to pattern recognition would use classification accuracy as payoff (utility) function, to give feedback to the Learning Automata. In general, this leads to the Vanishing Signal to Noise Ratio Problem described in Section~\ref{sec:vanishing}. This is due to the relatively small effect each single Learning Automaton has on overall classification accuracy, and due to the noise introduced by the random behaviour of the team as a whole. In the Tsetlin Machine game, on the other hand, each single automaton gets feedback directly from the value of its literal and the value of its clause, as laid out in Table~\ref{table:type_i_feedback} and Table~\ref{table:type_ii_feedback}. We will now show that this local feedback scheme can be a generalized ordinal potential game \cite{monderer1996potential}, with classification accuracy being the global payoff (utility) function.

\begin{mytheorem}
Let $M$ refer to the payoff matrix of the Tsetlin Machine game from Definition~\ref{def:tsetlin_machine_game}. Further, let $a_i'$ and $a_i''$ refer to the two actions available to a single automaton, while $a_{{-i}}$ refers to the remaining actions in the action configuration $\overline{\alpha}$. Finally, let the summation target $T$ be $1$, only considering positive clauses. The resulting Tsetlin Machine game is a generalized ordinal potential game with regards to classification accuracy, under the constraints of Definition \ref{def:problem_definition}. That is, we have: $P(y = \hat{y}; a'_{{i}},a_{{-i}}) - P(y = \hat{y}; a''_{{i}},a_{{-i}})>0\Rightarrow M (a'_{{i}},a_{{-i}}) - M(a''_{{i}},a_{{-i}})>0$.
\end{mytheorem}

\begin{proof}
 Classification accuracy is clearly maximised only when $C^{1}_{j}, j \in \{1, \ldots, n/2\},$ are duplicating $Q^{1}_{h}, h \in \{1, \ldots, n/2\}$. Any mismatch will either produce false positive output if the Tsetlin Machine clauses $C^{1}_{j}$ are missing literals, or false negative output if not all of the sub-patterns  $Q^{1}_{h}$ are completely covered. Thus, during learning, there are two ways classification accuracy can increase by switching action from $a''_i$ to $a'_i$: either by a decrease in the rate of false negatives, $P(y=1, \hat{y} = 0)$, or by a decrease in the rate of false positives, $P(y=0, \hat{y} = 1)$.

Let us first consider decreasing $P(y=1, \hat{y} = 0)$. This can only be achieved by turning false negative output into true positive output. Clearly, for a particular input $X$, we have false negative output if there exists an $h$ such that $Q^1_h(X) = 1$, with $C^1_j(X) = 0$ for all $j$. Let $L^1_j$ be the literals of clause $C^1_j$ and $K^1_h$ be the literals of $Q^1_h$. We can correct the false negative error \emph{if and only if} we have a $j$ and $k$ so that $L^{1}_{j} \setminus \{l_k\} \subseteq K^{1}_{h}, l_k \notin K^{1}_{h}$, implying $l_k=0$. The false negative error can then be corrected by excluding $l_k$ from clause $C^1_j$. Examining Feedback Type I for $C^1_j(X)=0, l_k=0,$ in Table \ref{table:type_i_feedback}, which applies for $y=1$, we observe that the expected payoff of excluding $l_k$ is larger that the expected payoff of including it.

We consider decreasing $P(y=0, \hat{y} = 1)$ next. This can be achieved by turning false positive output into true negative output. Clearly, for a particular input $X$, we have false positive output \emph{if and only if} there exists a $j$ such that $C^1_j(X) = 1$, with $Q^1_h(X) = 0$ for all $h$. Let $L^1_j$ be the literals of clause $C^1_j$ and $K^1_h$ be the literals of $Q^1_h$. We can correct the false positive output \emph{if and only if} there exists \emph{only one} $j$ such that $C^1_j(X) = 1$, and if this particular $C^1_j(X)$ has left out a literal $l_k$ of value $0$, $l_k = 0, l_k \notin L^1_j$. Then the false positive output can be corrected by including $l_k=0$ in $C^1_j$. Examining Feedback Type II for $C^1_j(X)=1, l_k=0,$ in Table \ref{table:type_ii_feedback}, which applies for $y=0$, we observe that the expected payoff of including $l_k$ is larger that the expected payoff of excluding it.
\end{proof}
\paragraph{Remark 4.} The ordinal potential game is \emph{generalized} because the Tsetlin Machine game matrix $M$ rewards action switches that have no impact on classification accuracy. That is, we do not have:
$M(a'_{{i}},a_{{-i}}) - M(a''_{{i}},a_{{-i}}) > 0 \Rightarrow P(y = \hat{y}; a'_{{i}},a_{{-i}}) - P(y = \hat{y}; a''_{{i}},a_{{-i}})>0$. In particular, consider false negative output that cannot be corrected by excluding a single literal from a single clause. Yet, $M$ still rewards excluding literals of value $0, l_k=0,$ when $C^1_j(X)=0$, \emph{eventually} correcting the false negative output. Similarly, when multiple clauses output $1$, producing a false positive output, no single \emph{Include} action can alone correct the false positive output. Yet, including literals of value $0$ is still rewarded in the game matrix $M$, again, \emph{eventually} fixing the false positive output. Thus, in this sense, the Tsetlin Machine game matrix has a longer planning horizon than just greedily optimizing classification accuracy in single steps. Note that these latter dynamics do not produce any additional Nash equilibria, as established by Theorem \ref{theorem:optimality_without_noise_system}. This means that there is a one-to-one correspondence between the equilibria in the potential game and in optimizing classification accuracy.

\paragraph{Remark 5.} As summarized in \cite{Narendra1989}, the behaviour of finite state Learning Automata collectives have been studied by several researchers, showing that rational behaviour can be obtained in certain cases, when memory is infinite and the game's payoff matrix is constrained \cite{Krylov1963Games,Tung1996,Borovikov1965,Volkonskii1965,Pittel1965}. Proofs for variable structure Learning Automata are more general, on the other hand. That is, for identical payoff games, the Learning Automata collective converges to a Nash equilibrium with probability arbitrarily close to unity \cite{Narendra1989}. Further, for zero-sum games with a saddle point, there exists Learning Automata collectives that can obtain the Von Neumann value \cite{Viswanathan74,Narendra1974competitive}. In the next section, we study the behaviour of Tsetlin Machine games empirically on various datasets, including the artificial Noisy XOR dataset where the optimal classification accuracy and the corresponding Nash equilibria are known. 

\section{Empirical Results}
\label{sec:empirical_results}

In this section, we evaluate the Tsetlin Machine empirically using five datasets:
\begin{itemize}
    \item {\bf Binary Iris Dataset.} This is the classical Iris Dataset, however, with features in binary form.
    \item {\bf Binary Digits Dataset.} This is the classical digits dataset, again with features in binary form.
    \item {\bf \emph{Axis \& Allies} Board Game Dataset.} This new dataset involves optimal move prediction in a minimalistic, yet intricate, mini-game from the \emph{Axis \& Allies} board game.
    \item {\bf Noisy XOR Dataset with Non-informative Features.} This artificial dataset is designed to reveal particular "blind zones" of pattern recognition algorithms. The dataset captures the renowned XOR-relation. Furthermore, the dataset contains a large number of random non-informative features to measure susceptibility towards the curse of dimensionality \cite{Duda2000}. To examine robustness towards noise we have further randomly inverted $40\%$ of the outputs.
    \item {\bf MNIST Dataset.} The MNIST dataset is a larger scale dataset used extensively to benchmark machine learning algorithms. We have included this dataset to investigate the scalability of the Tsetlin Machine, as well as the behaviour of longer learning processes.
\end{itemize}

For these datasets, we form ensembles of $50$ to $1000$ independent replications with different random number streams. We do this to minimize the variance of the reported results and to provide the foundation for a statistical analysis of the merits of the different schemes evaluated. 

\begin{figure}[!ht]
\centering
\includegraphics[width=6.0in]{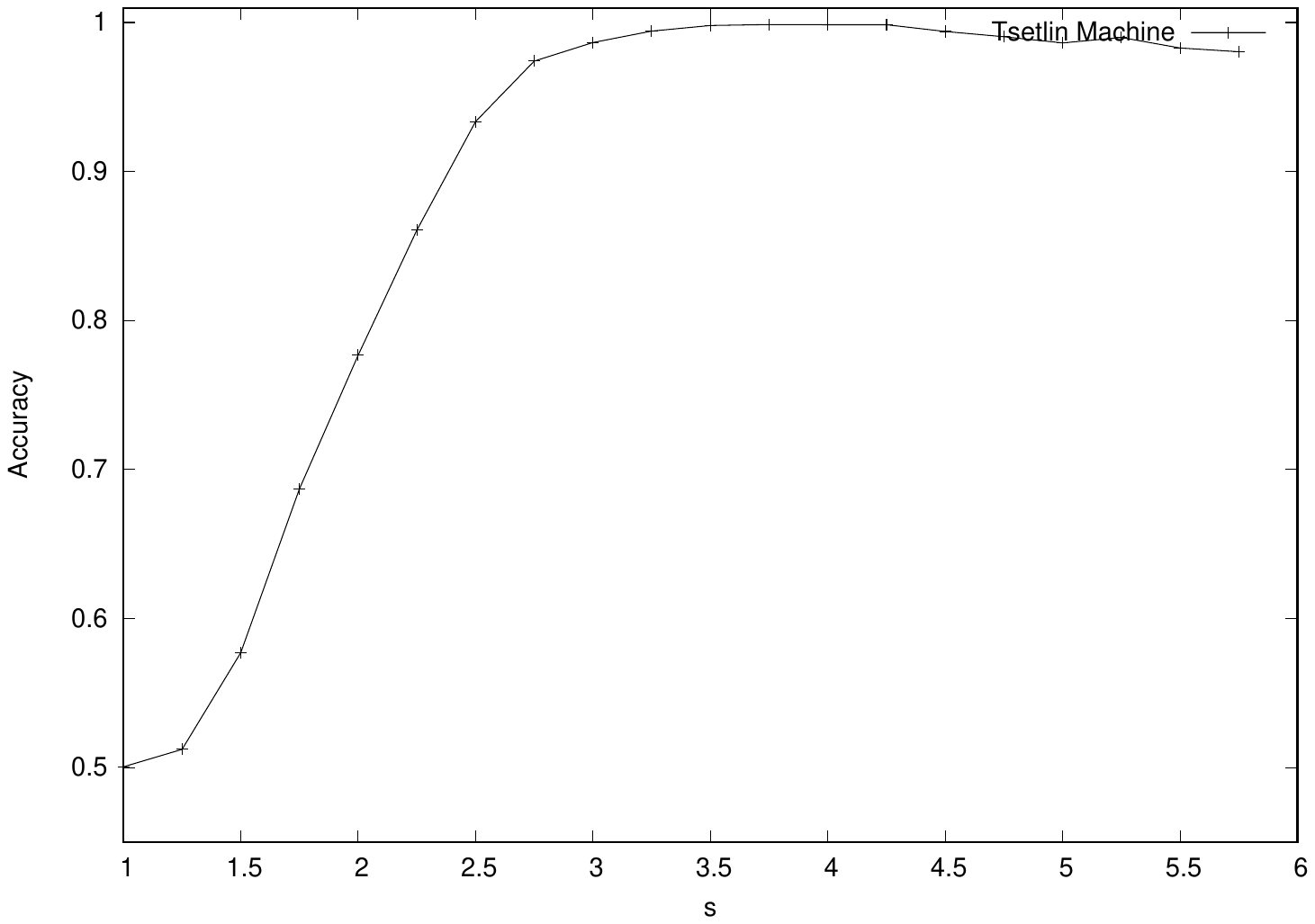}
\caption{The mean accuracy of the Tsetlin Machine (y-axis) on the Noisy XOR Dataset for different values of the parameter $s$ (x-axis).}
\label{figure:xor_s}
\end{figure}

Together with the Tsetlin Machine, we also evaluate several classical machine learning techniques using the same random number streams. This  includes Multilayer Perceptron Networks, the Naive Bayes Classifier, Support Vector Machines, and Logistic Regression. Where appropriate, the different schemes are optimized by means of relatively light hyper-parameter grid searches.
As an example, Figure \ref{figure:xor_s} captures the impact the $s$ parameter of the Tsetlin Machine has on mean accuracy, for the Noisy XOR Dataset. Each point in the plot measures the mean accuracy of $100$ different replications of the XOR-experiment for a particular value of $s$. Clearly, accuracy increases with $s$ up to a certain point, before it degrades gradually. Based on the plot, for the Noisy XOR-experiment, we decided to use an $s$ value of $3.9$.

\subsection{The Binary Iris Dataset}

We first evaluate the Tsetlin Machine on the classical Iris dataset\footnote{UCI Machine Learning Repository [\href{https://archive.ics.uci.edu/ml/datasets/iris}{https://archive.ics.uci.edu/ml/datasets/iris}].}. This dataset consists of 150 examples with four inputs (Sepal Length, Sepal Width, Petal Length and Petal Width), and three possible outputs (Setosa, Versicolour, and Virginica).

We increase the challenge by transforming the four input values into one consecutive sequence of $16$ bits, four bits per float. It is thus necessary to also learn how to segment the $16$ bits into four partitions, and extract the numeric information. We refer to the new dataset as the Binary Iris Dataset.

We partition this dataset into a training set and a test set, with 80 percent of the data being used for training. We here randomly produce $1000$ training and test data partitions. For each ensemble, we also randomly reinitialize the competing algorithms, to gain information on stability and robustness. The results are reported in Table \ref{tab:accuracy_binary_iris_test}.

\begin{table}[!bh]
    \centering
    \begin{tabular}{r||c|c|c|c|c}
         \bf Technique/Accuracy ($\%$)&\bf Mean&\bf $5~\%$ile &\bf $95~\%$ile&\bf Min.&\bf Max.\\
         \hline
    Tsetlin Machine&$95.0 \pm 0.2$&$86.7$&$100.0$&$80.0$&$100.0$\\
    Naive Bayes&$91.6 \pm 0.3$&$83.3$&$96.7$&$70.0$&$100.0$\\
    Logistic Regression&$92.6 \pm 0.2$&$86.7$&$100.0$&$76.7$&$100.0$\\
    Multilayer Perceptron Networks&$93.8 \pm 0.2$&$86.7$&$100.0$&$80.0$&$100.0$\\
    SVM&$93.6 \pm 0.3$&$86.7$&$100.0$&$76.7$&$100.0$
    \end{tabular}
    \caption{The Binary Iris Dataset -- accuracy on test data.}
    \label{tab:accuracy_binary_iris_test}
\end{table}

The Tsetlin Machine\footnote{In this experiment, we use a Multi-Class Tsetlin Machine, described in Section 6.1. We also apply Boosting of True  Positive  Feedback  to  \emph{Include} actions as described in Section \ref{sec:type_i_feedback}.} used here employs $300$ clauses, and uses an $s$-value of $3.0$ and a summation target $T$ of $10$. Furthermore, the individual Tsetlin Automata each has $100$ states. This Tsetlin Machine is run for $500$ epochs, and it is the accuracy after the final epoch that is reported. 
Propositional formulas with higher test accuracy are often found in preceding epochs because of the random exploration of the Tsetlin Machine. However, to avoid overfitting to the test set by handpicking the best configuration found, we instead simply use the last configuration produced.

In Table \ref{tab:accuracy_binary_iris_test}, we list mean accuracy with $95\%$ confidence intervals, $5$ and $95$ percentiles, as well as the minimum and maximum accuracy obtained, across the $1000$ experiment runs we executed. As seen, the Tsetlin Machine provides the highest mean accuracy. For the $95~\%$ile scores, however, most of the schemes obtain $100\%$ accuracy. This can be explained by the small size of the test set, which merely contains 30 examples. Thus it is easier to stumble upon a random configuration that happens to provide fault-free classification. Since the test set is merely a sample of the corresponding real-world problem, it is reasonable to assume that higher mean accuracy translates to more robust performance overall.

The training set, on the other hand, reveals subtler differences between the schemes. The results obtained on the training set are shown in Table \ref{tab:accuracy_binary_iris_training}.  As seen, the SVM here provides the highest mean accuracy, while the Tsetlin Machine provides the second highest. However, the large drop in accuracy from the training data to the test data for the SVM indicates overfitting on the training data.

\begin{table}[!bh]
    \centering
    \begin{tabular}{r||c|c|c|c|c}
         \bf Technique&\bf Mean&\bf $5~\%$ile &\bf $95~\%$ile&\bf Min.&\bf Max.\\
         \hline
    Tsetlin Machine&$96.6 \pm 0.05$&$95.0$&$98.3$&$94.2$&$99.2$\\
    Naive Bayes&$92.4 \pm 0.08$&$90.0$&$94.2$&$85.8$&$97.5$\\
    Logistic Regression&$93.8 \pm 0.07$&$92.5$&$95.8$&$90.0$&$97.5$\\
    Multilayer Perceptron Network&$95.0 \pm 0.07$&$93.3$&$96.7$&$92.5$&$98.3$\\
    SVM&$96.7 \pm 0.05$&$95.8$&$98.3$&$95.8$&$99.2$
    \end{tabular}
    \caption{The Binary Iris Dataset -- accuracy on training data.}
    \label{tab:accuracy_binary_iris_training}
\end{table}

\subsection{The Binary Digits Dataset}

We next evaluate the Tsetlin Machine on the classical Pen-Based Recognition of Handwritten Digits Dataset\footnote{UCI Machine Learning Repository [\href{http://archive.ics.uci.edu/ml/datasets/Pen-Based+Recognition+of+Handwritten+Digits}{http://archive.ics.uci.edu/ml/datasets/Pen-Based+Recognition+of+Handwritten+Digits}].}. The original dataset consists of 250 handwritten digits from 44 different writers, for a total number of $10 992$ instances. We increase the challenge by removing the individual pixel value structure, transforming the $64$ different input features into a sequence of $192$ bits, $3$ bits per pixel. We refer to the modified dataset as the Binary Digits Dataset. Again we partition the dataset into training and test sets, keeping 80 percent of the data for training.

The Tsetlin Machine\footnote{In this experiment, we used a Multi-Class Tsetlin Machine, described in Section 6.1.  We also apply Boosting of True  Positive  Feedback  to  \emph{Include} actions as described in Section \ref{sec:type_i_feedback}.} used here contains $1000$ clauses, and uses a specificity $s$ of $3.0$ and a summation target $T$ of $10$. Furthermore, the individual Tsetlin Automata each has $1000$ states. The Tsetlin Machine is run for $300$ epochs, and it is the accuracy after the final epoch that is reported.

Table \ref{tab:accuracy_binary_digits_test} reports mean accuracy with $95\%$ confidence intervals, $5$ and $95$ percentiles, as well as the minimum and maximum accuracy obtained, across the $100$ experiment runs we executed. As seen, the Tsetlin Machine again clearly provides the highest accuracy on average, also when taking the $95\%$ confidence intervals into account. For this dataset, the Tsetlin Machine is also superior when it comes to the maximal accuracy found across the $100$ replications of the experiment, as well as for the $95~\%$ile results.

\begin{table}[!bh]
    \centering
    \begin{tabular}{r||c|c|c|c|c}
         \bf Technique/Accuracy ($\%$)&\bf Mean&\bf $5~\%$ile &\bf $95~\%$ile&\bf Min.&\bf Max.\\
         \hline
    Tsetlin Machine&$95.7 \pm 0.2$&$93.9$&$97.2$&$92.5$&$98.1$\\
    Naive Bayes&$91.3 \pm 0.3$&$88.9$&$93.6$&$87.2$&$94.4$\\
    Logistic Regression&$94.0 \pm 0.2$&$91.9$&$95.8$&$90.8$&$96.9$\\
    Multilayer Perceptron Network&$93.5 \pm 0.2$&$91.7$&$95.3$&$90.6$&$96.7$\\
    SVM&$50.5 \pm 2.2$&$30.3$&$67.4$&$25.8$&$77.8$
    \end{tabular}
    \caption{The Binary Digits Dataset -- accuracy on test data.}
    \label{tab:accuracy_binary_digits_test}
\end{table}

Performing poor on the test data and well on the training data indicates susceptibility to overfitting. Table \ref{tab:accuracy_binary_digits_training} reveals that the other techniques, apart from the Naive Bayes Classifier, perform significantly better on the training data, unable to transfer this performance to the test data.

Table \ref{tab:bit_pattern} visualizes one of the clauses produced by the Tsetlin Machine, capturing handwritten digit '1'. As seen, the pattern should be relatively easy to interpret for humans compared to, e.g., a neural network.

\begin{table}[!bh]
    \centering
    \begin{tabular}{r||c|c|c|c|c}
         \bf Technique&\bf Mean&\bf $5~\%$ile &\bf $95~\%$ile&\bf Min.&\bf Max.\\
         \hline
    Tsetlin Machine&$100.0 \pm 0.01$&$99.9$&$100.0$&$99.8$&$100.0$\\
    Naive Bayes&$92.9 \pm 0.07$&$92.4$&$93.5$&$91.3$&$93.7$\\
    Logistic Regression&$99.6 \pm 0.02$&$99.4$&$99.7$&$99.3$&$99.9$\\
    Multilayer Perceptron Network&$100.0 \pm 0.0$&$100.0$&$100.0$&$100.0$&$100.0$\\
    SVM&$100.0 \pm 0.0$&$100.0$&$100.0$&$100.0$&$100.0$
    \end{tabular}
    \caption{The Binary Digits Dataset -- accuracy on training data.}
    \label{tab:accuracy_binary_digits_training}
\end{table}

\subsection{The \emph{Axis \& Allies} Board Game Dataset}

Besides the two classical datasets, we also have built a new dataset based on the board game \emph{Axis \& Allies}\footnote{\href{http://avalonhill.wizards.com/games/axis-and-allies}{http://avalonhill.wizards.com/games/axis-and-allies}}. 
We designed this dataset to exhibit intricate pattern structures, involving optimal move prediction in a subgame of \emph{Axis \& Allies}. In \emph{Axis \& Allies}, every piece on the board are potentially moved each turn. Additionally, new pieces are introduced throughout the game, as a result of earlier decisions. This arguably yields a larger search tree than the ones we find in Go and chess. Finally, the outcome of battles are determined by dice, rendering the game stochastic.

\begin{figure}[!th]
\centering
\includegraphics[width=3.0in]{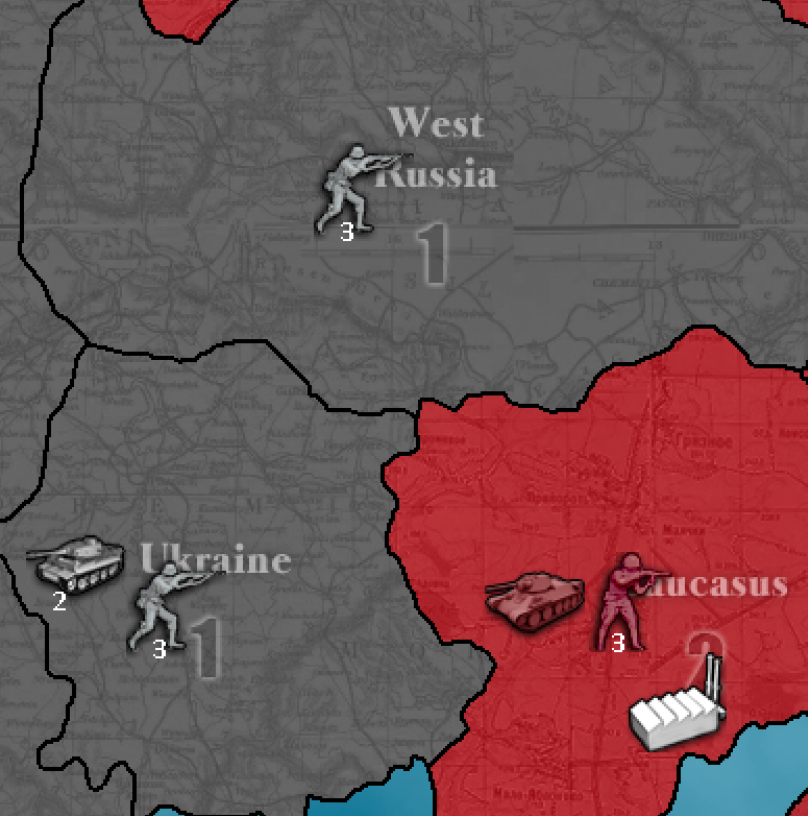}
\caption{The \emph{Axis \& Allies} mini game.}
\label{figure:axis_and_allies_minigame}
\end{figure}

The \emph{Axis \& Allies} Board Game Dataset consists of $10~000$ board game positions, exemplified in Figure \ref{figure:axis_and_allies_minigame}. Player 1 owns the "Caucasus" territory in the figure, while Player 2 owns "Ukraine" and "West Russia". At start-up, each player is randomly assigned 0-10 tanks and 0-20 infantry each. These units are the starting forces. For Player 2, the units are randomly distributed among his two territories. The game consists of two rounds. First Player 1 attacks. This is followed by a counter attack by Player 2. In order to win, Player 1 needs to capture both of "Ukraine" and "West Russia". Player 2, on the other hand, only needs to take "Caucasus".

\begin{table}[!bh]
    \centering
    \begin{tabular}{c|c|c|c|c|c||c|c|c|c}
    \multicolumn{6}{c||}{\bf At Start}&\multicolumn{4}{c}{\bf Optimal Attack}\\
    \hline
    \multicolumn{2}{c|}{\bf Caucasus}&\multicolumn{2}{c|}{\bf W. Russia}&\multicolumn{2}{c||}{\bf Ukraine}&\multicolumn{2}{c|}{\bf W. Russia}&\multicolumn{2}{c}{\bf Ukraine}\\
     \hline
    \bf Inf&\bf Tnk&\bf Inf&\bf Tnk&\bf Inf&\bf Tnk&\bf Inf&\bf Tnk&\bf Inf&\bf Tnk\\
    \hline
    \hline
    16&4&11&4&5&4&0&0&3&4\\
    19&3&6&1&6&3&7&2&12&1\\
    9&1&1&3&0&5&0&0&0&0
    \end{tabular}
    \caption{The \emph{Axis \& Allies} Board Game Dataset.}
    \label{table:aa_dataset}
\end{table}

To produce the dataset, we built an \emph{Axis \& Allies} Board Game  simulator. This allowed us to find the optimal attack for each assignment of starting forces. The resulting input and output variables are shown in Table \ref{table:aa_dataset}. The at start forces are to the left, while the optimal attack forces can be found to the right. In the first row, for instance, it is optimal for Player 1 to launch a preemptive strike against the armor in Ukraine (armor is better offensively than defensively), to destroy offensive power, while keeping the majority of forces for defense.

We use $25\%$ of the data for training, and $75\%$ for testing, randomly producing $100$ different partitions of the dataset. The Tsetlin Machine employed here contains $10~000$ clauses, and uses an $s$-value of $40.0$ and a summation target $T$ of $10$. Furthermore, the individual Tsetlin Automata each has $1000$ states. The Tsetlin machine is run for $200$ epochs, and it is the accuracy after the final epoch that is reported.

Table \ref{tab:aa_test} reports the results from predicting output bit $5$ among the $20$ output bits (as representative for all of the bits). In the table, we list mean accuracy with $95\%$ confidence intervals, $5$ and $95$ percentiles, as well as the minimum and maximum accuracy obtained, across the $100$ experiment runs we executed. 
\begin{table}[!bh]
    \centering
    \begin{tabular}{r||c|c|c|c|c}
         \bf Technique/Accuracy ($\%$)&\bf Mean&\bf $5~\%$ile &\bf $95~\%$ile&\bf Min.&\bf Max.\\
         \hline
    Tsetlin Machine&$87.7 \pm 0.0$&$87.4$&$88.0$&$87.2$&$88.1$\\
    Naive Bayes&$80.1 \pm 0.0$&$80.1$&$80.1$&$80.1$&$80.1$\\
    Logistic Regression&$77.7 \pm 0.0$&$77.7$&$77.7$&$77.7$&$77.7$\\
    Multilayer Perceptron Network&$87.6 \pm 0.1$&$87.1$&$88.1$&$86.6$&$88.3$\\
    SVM&$83.7 \pm 0.0$&$83.7$&$83.7$&$83.7$&$83.7$\\
    Random Forest&$83.1 \pm 0.1$&$82.3$&$83.8$&$81.6$&$84.1$
    \end{tabular}
    \caption{The \emph{Axis \& Allies} Dataset -- accuracy on test data.}
    \label{tab:aa_test}
\end{table}
As seen in the table, apparently only the Tsetlin Machine and the neural network are capable of properly handling the complexity of the dataset, providing statistically similar performance. The Tsetlin Machine is quite stable performance-wise, while the neural network performance varies more.

However, the number of clauses needed to achieve the above performance is quite high for the Tsetlin Machine, arguably due to its flat one-layer architecture. Another reason that can explain the need for a large number of clauses can be the intricate nature of the mini-game of Axis \& Allies. Since we need an $s$-value as large as $40$, clearly, some of the pertinent sub-patterns must be quite fine-grained. Because the $s$-value is global, all patterns, even the coarser ones, must be learned at this fine granularity. A possible next step in the research on the Tsetlin Machine could therefore be to investigate the effect of having clauses with different $s$-values -- some with smaller values for the rougher patterns, and some with larger values for the finer patterns.

As a final observation, Table \ref{tab:aa_training} reports performance on the training data. Random Forest distinguishes itself by almost perfect predictions for the training data, thus clearly overfitting, but still performing well on the test set. The other techniques provide slightly improved performance on the training data, as expected.

\begin{table}[!bh]
    \centering
    \begin{tabular}{r||c|c|c|c|c}
         \bf Technique/Accuracy ($\%$)&\bf Mean&\bf $5~\%$ile &\bf $95~\%$ile&\bf Min.&\bf Max.\\
         \hline
    Tsetlin Machine&$96.2 \pm 0.1$&$95.7$&$96.8$&$95.5$&$97.0$\\
    Naive Bayes&$81.2 \pm 0.0$&$81.2$&$81.2$&$81.2$&$81.2$\\
    Logistic Regression&$78.8 \pm 0.0$&$78.8$&$78.8$&$78.8$&$78.8$\\
    Multilayer Perceptron Network&$92.6 \pm 0.1$&$91.5$&$93.6$&$90.7$&$94.2$\\
    SVM&$85.2 \pm 0.0$&$85.2$&$85.2$&$85.2$&$85.2$\\
    Random Forest&$99.1 \pm 0.0$&$98.8$&$99.4$&$98.6$&$99.7$
    \end{tabular}
    \caption{The \emph{Axis \& Allies} Dataset -- accuracy on training data.}
    \label{tab:aa_training}
\end{table}

\subsection{The Noisy XOR Dataset with Non-informative Features}

We now turn to an artifical dataset, constructed to uncover "blind zones" caused by XOR-like relations. Furthermore, the dataset contains a large number of random non-informative features to measure susceptibility towards the curse of dimensionality \cite{Duda2000}. To examine robustness towards noise, we have further randomly inverted $40\%$ of the outputs.

\begin{table}[!bh]
    \centering
    \begin{tabular}{c|c|c|c|c|c|c|c|c|c|c|c||c}
    \bf $x_1$&\bf $x_2$&\bf $x_3$&\bf $x_4$&\bf $x_5$&\bf $x_6$&\bf $x_7$&\bf $x_8$&\bf $x_9$&\bf $x_{10}$&\bf $x_{11}$&\bf $x_{12}$&\bf $y$\\
    \hline
    \hline
    0&1&0&1&1&0&1&1&0&1&1&0&1\\
    1&1&1&0&1&0&1&1&0&0&1&1&0\\
    0&0&1&1&0&1&1&1&1&0&1&0&0\\
    1&1&1&0&1&1&1&0&1&1&0&0&1
    \end{tabular}
    \caption{The Noisy XOR Dataset with Non-informative Features.}
    \label{tab:noisy_xor_dataset}
\end{table}

\begin{table}[!bh]
    \centering
    \begin{tabular}{c|c|c}
    {\bf No.}&{\bf Sign}&{\bf Clause Learned}\\
    \hline 
    \hline
    1&$+$&$\lnot x_1 \land x_2$\\
    2&$-$&$\lnot x_1 \land \lnot x_2$\\
    3&$+$&$x_1 \land \lnot x_2$\\
    4&$-$&$x_1 \land x_2$
    \end{tabular}
    \caption{Example of four clauses composed by the Tsetlin Machine for the XOR Dataset with Non-informative Features.}
    \label{tab:noisy_xor_dataset_clauses}
\end{table}

The dataset consists of $10~000$ examples with twelve binary inputs, $X = [x_1, x_2, \ldots, x_{12}]$, and a binary output, $y$. Ten of the inputs are completely random. The two remaining inputs, however, are related to the output $y$ through an XOR-relation, $y = \mathrm{XOR}(x_{k_1}, x_{k_2})$. Finally, $40\%$ of the outputs are inverted. Table \ref{tab:noisy_xor_dataset} shows four examples from the dataset, demonstrating the high level of noise. We partition the dataset into training and test data, using $50\%$ of the data for training.

The Tsetlin Machine\footnote{In this experiment, we used a Multi-Class Tsetlin Machine, described in Section 6.1.} used here contains $20$ clauses, and uses an $s$-value of $3.9$ and a summation target $T$ of $15$. Furthermore, the individual Tsetlin Automata each has $100$ states. The Tsetlin Machine is run for $200$ epochs, and it is the accuracy after the final epoch, that we report.

Table \ref{tab:noisy_xor_dataset_clauses} contains four of the clauses produced by the Tsetlin Machine. Notice how the noisy dataset from Table \ref{tab:noisy_xor_dataset} has been turned into informative propositional formulas that capture the structure of the dataset.

The empirical results are found in Table \ref{tab:xor_test}. Again, we report mean accuracy with $95\%$ confidence intervals, $5$ and $95$ percentiles, as well as the minimum and maximum accuracy obtained, across the $100$ replications of the experiment. Note that for the test data, the output values are unperturbed. As seen, the XOR-relation, as expected, makes Logistic Regression and the Naive Bayes Classifier incapable of predicting the output value $y$, resorting to random guessing. Both the neural network and the Tsetlin Machine, on the other hand, see through the noise and captures the underlying XOR pattern. SVM performs slightly better than the Naive Bayes Classifier and Logistic Regression, however, is clearly distracted by the added non-informative features (the SVM performs much better with fewer non-informative features).

\begin{table}[!bh]
    \centering
    \begin{tabular}{r||c|c|c|c|c}
         \bf Technique/Accuracy ($\%$)&\bf Mean&\bf $5~\%$ile &\bf $95~\%$ile&\bf Min.&\bf Max.\\
         \hline
    Tsetlin Machine&$99.3 \pm 0.3$&$95.9$&$100.0$&$91.6$&$100.0$\\
    Naive Bayes&$49.8 \pm 0.2$&$48.3$&$51.0$&$41.3$&$52.7$\\
    Logistic Regression&$49.8 \pm 0.3$&$47.8$&$51.1$&$41.1$&$53.1$\\
    Multilayer Perceptron Network&$95.4 \pm 0.5$&$90.1$&$98.6$&$88.2$&$99.9$\\
    SVM&$58.0 \pm 0.3$&$56.4$&$59.2$&$55.4$&$66.5$
    \end{tabular}
    \caption{The Noisy XOR Dataset with Non-informative Features -- accuracy on test data.}
    \label{tab:xor_test}
\end{table}

\begin{figure}[!ht]
\centering
\includegraphics[width=6.0in]{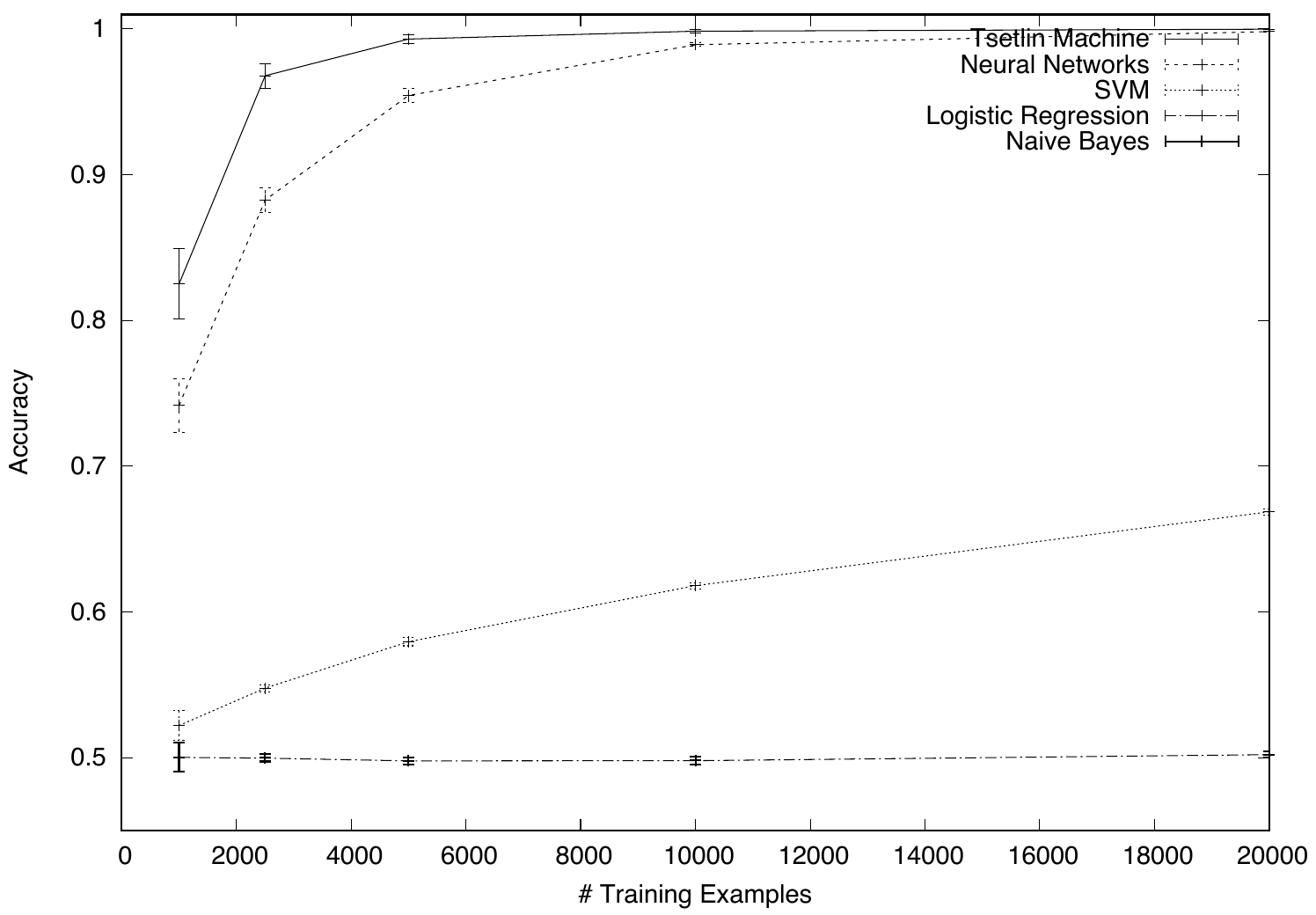}
\caption{Accuracy (y-axis) for the Noisy XOR Dataset for different training dataset sizes (x-axis).}
\label{figure:xor_data_sizes}
\end{figure}

Figure \ref{figure:xor_data_sizes} shows how accuracy degrades with less data, when we vary the dataset size from $1000$ examples to $20~000$ examples. As expected, Naive Bayes and Logistic Regression guess blindly for all the different data sizes. The main observation, however, is that the accuracy advantage the Tsetlin Machine has over  neural networks increases with less training data. Indeed, it turns out that the Tsetlin Machine performs robustly with small training data sets in all of our experiments.

\subsection{The MNIST Dataset}

\begin{figure}[!ht]
\centering
\includegraphics[width=6.0in]{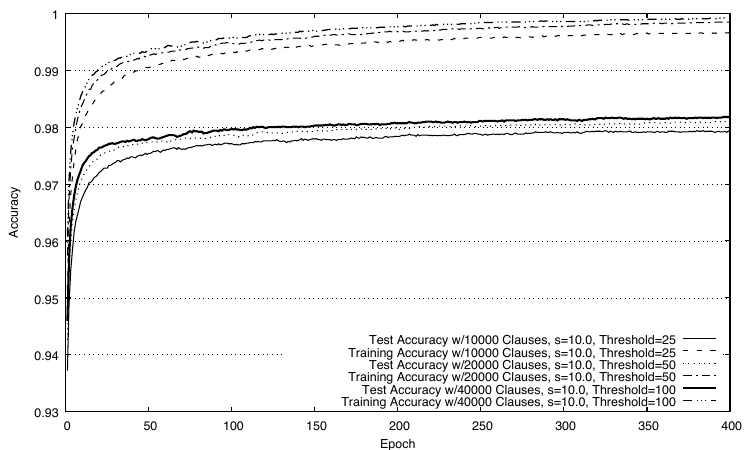}
\caption{The mean test- and training accuracy per epoch for the Tsetlin Machine on the MNIST Dataset.}
\label{figure:learning_progress}
\end{figure}

We next evaluate the Tsetlin Machine on the MNIST Dataset of Handwritten Digits\footnote{\href{http://www.pymvpa.org/datadb/mnist.html}{http://www.pymvpa.org/datadb/mnist.html}} \cite{LeCun1998}, also investigating how learning progresses, epoch-by-epoch, in terms of accuracy. Note that the experimental results reported here can be reproduced with the demo found at\\ \href{ https://github.com/cair/fast-tsetlin-machine-with-mnist-demo}{https://github.com/cair/fast-tsetlin-machine-with-mnist-demo}.

The original dataset consists of $60~000$ training examples, and $10~000$ test examples. We binarize this dataset by replacing pixel values larger than $0.3$ with $1$. Pixel values below or equal to $0.3$ are replaced with $0$. 

The Tsetlin Machine\footnote{In this experiment, we used a Multi-Class Tsetlin Machine, described in Section 6.1.  We also applied Boosting of True  Positive  Feedback  to  \emph{Include} actions, as described in Section \ref{sec:type_i_feedback}.} employed here contains $40~000$ clauses, $4000$ clauses per class, uses an $s$-value of $10.0$, and a summation target $T$ of $50$. Furthermore, the individual Tsetlin Automata each has $256$ states. The Tsetlin Machine is run for $400$ epochs, and it is the accuracy after the final epoch that is reported.

As seen in Figure \ref{figure:learning_progress}, both mean test- and training accuracy increase almost monotonically across the epochs, however, affected by random fluctuation. Perhaps most notably, while the mean accuracy on the training data approaches 
$99.9$\%, accuracy on the test data continues to increase as well, hitting $98.2$\% after 400 epochs. This is quite different from what occurs with back-propagation on a neural network, where accuracy on test data starts to drop at some point due to overfitting, without proper regularization mechanisms.

The figure also shows how varying the number of clauses and the summation target $T$ affects accuracy and learning stability. With more clauses available to express patterns, in combination with a higher summation target $T$, both learning speed, stability and accuracy increases, however, at the expense of larger computational cost.

\begin{table}[!bh]
    \centering
    \begin{tabular}{r||c}
        \bf Technique&\bf Accuracy ($\%$)\\
        \hline
        {\it 2-layer NN, 800 HU, Cross-Entropy Loss}&$\mathit{98.6}$\\
        Tsetlin Machine ($95$ \%ile)&$98.3$\\
        Tsetlin Machine (Mean)&$98.2 \pm 0.0$\\
        Tsetlin Machine ($5$ \%ile)&$98.1$\\
        {\it K-nearest-neighbors, L3}&$\mathit{97.2}$\\
        {\it 3-layer NN, 500+150 hidden units}&$\mathit{97.1}$\\
        {\it 40 PCA + quadratic classifier}&$\mathit{96.7}$\\
        {\it 1000 RBF + linear classifier}&$\mathit{96.4}$\\
        Logistic regression&$91.5$\\
        {\it Linear classifier (1-layer NN)}&$\mathit{88.0}$\\
        Decision tree&$87.8$\\
        Multinomial Naive Bayes&$83.2$
    \end{tabular}
    \caption{A comparison of vanilla machine learning algorithms with the Tsetlin Machine, directly on the original unenhanced MNIST dataset (NN - Neural Network). }
    \label{tab:MNIST_comparison}
\end{table}

Table \ref{tab:MNIST_comparison} reports the mean accuracy of the Tsetlin Machine, across the $50$ experiment runs we executed. As points of reference, results for other well-known algorithms have been obtained from \href{http://yann.lecun.com/exdb/mnist/}{http://yann.lecun.com/exdb/mnist/} and included in the table (in italic). Only the vanilla version of the algorithms, that has been applied directly on unenhanced MNIST data, is included here. The purpose of this selection is to strictly compare algorithmic performance. In other words, we do not consider the effect of enhancing the dataset (e.g., warping, distortion, deskewing), combining different algorithms (e.g., neural network combined with nearest neighbor, convolution schemes), or applying meta optimization techniques (boosting, ensemble learning, etc.). With such techniques, it is possible to significantly increase accuracy, with the best currently reported results being an accuracy of $99.79$\% \cite{Wan2013}. Enhancing the vanilla Tsetlin Machine with such techniques is further work.

Additionally, as a further point of reference, we train and evaluate logistic regression, decision trees, and multinomial Naive Bayes on the \emph{binarized} MNIST dataset used by the Tsetlin Machine.

As seen in the table, the Tsetlin Machine provides competitive accuracy, outperforming e.g. K-nearest neighbor and a 3-layer neural network. It is outperformed by a 2-layer neural network with 800 hidden nodes, using cross entropy loss. However, note that the Tsetlin Machine operates upon the binarized MNIST data (the grey tone value of each pixel is either set to 0 or 1), and thus has a disadvantage. Improved binarization techniques for the Tsetlin Machine is further work.

\section{The Tsetlin Machine as a Building Block in More Advanced Architectures}
\label{sec:building_block}

We have designed the Tsetlin Machine to facilitate building of more advanced architectures. We will here exemplify different ways of connecting multiple Tsetlin Machines in more advanced architectures.

\subsection{The Multi-Class Tsetlin Machine}

In some pattern recognition problems the task is to assign one of $n$ classes to each observed pattern $X$. That is, one needs to decide upon a \emph{single} output value, $y \in \{1, \ldots, m\}$. Such a multi-class pattern recognition problem can be handled by the Tsetlin Machine by representing $y$ as bits, using multiple outputs $y$. In this section, however, we present an alternative architecture that addresses the multi-class pattern recognition problem more directly.

\begin{figure}[!th]
\centering
\includegraphics[width=4.5in]{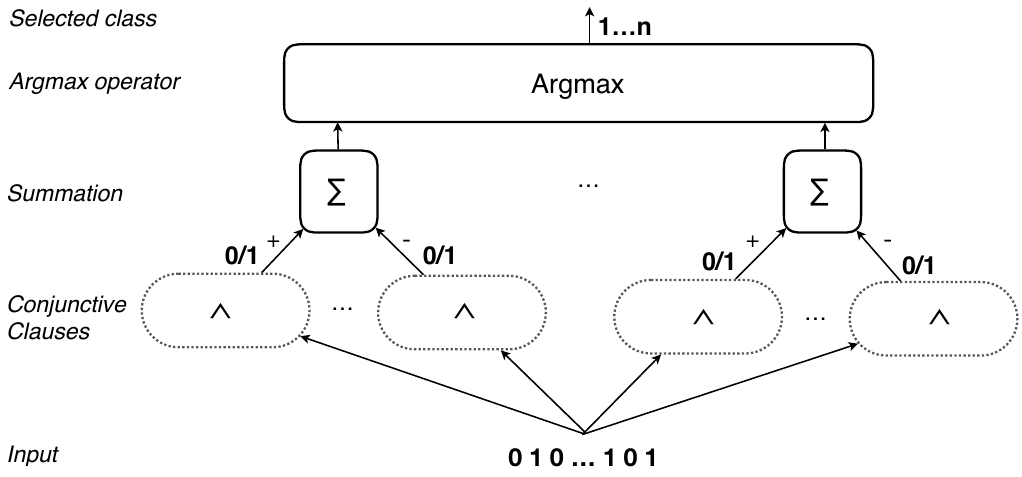}
\caption{The Multi-Class Tsetlin Machine.}
\label{figure:multi-class_tsetlin_machine}
\end{figure}

Figure \ref{figure:multi-class_tsetlin_machine} depicts the Multi-Class Tsetlin Machine\footnote{An implementation of the Multi-Class Tsetlin Machine can be found at \href{https://github.com/cair/TsetlinMachine}{https://github.com/cair/TsetlinMachine}.} which replaces the threshold function of each output $y, i \in \{1, \ldots, m\}$, with a single {\bf argmax} operator. With the {\bf argmax} operator, the index $i$ of the largest sum $\sum_{j=1}^{n/2} C_j^{1,i}(X) -\sum_{j=1}^{n/2} C_j^{0,i}(X)$ is outputted as the final output of the Tsetlin Machine:
\begin{equation}
y = \mathrm{argmax}_{i=1,\ldots,m}\left(\sum_{j=1}^{n/2} C_j^{1,i}(X) -\sum_{j=1}^{n/2} C_j^{0,i}(X)\right).
\end{equation}

Training is done as described in Section \ref{sec:tsetlin_machine}, apart from one critical modification. Assume we have $y=i$ for the current observation $(X,y)$. Then the Tsetlin Automata teams associated with class $i$ are trained as per $y=1$ in the original Algorithm \ref{alg:tsetlin_machine}. Additionally, a random class $q \ne i$ is selected. The Tsetlin Automata teams of class $q$ are then trained in accordance with $y=0$ in the original algorithm.

\subsection{The Fully Connected Deep Tsetlin Machine}

\begin{figure}[!th]
\centering
\includegraphics[width=4.0in]{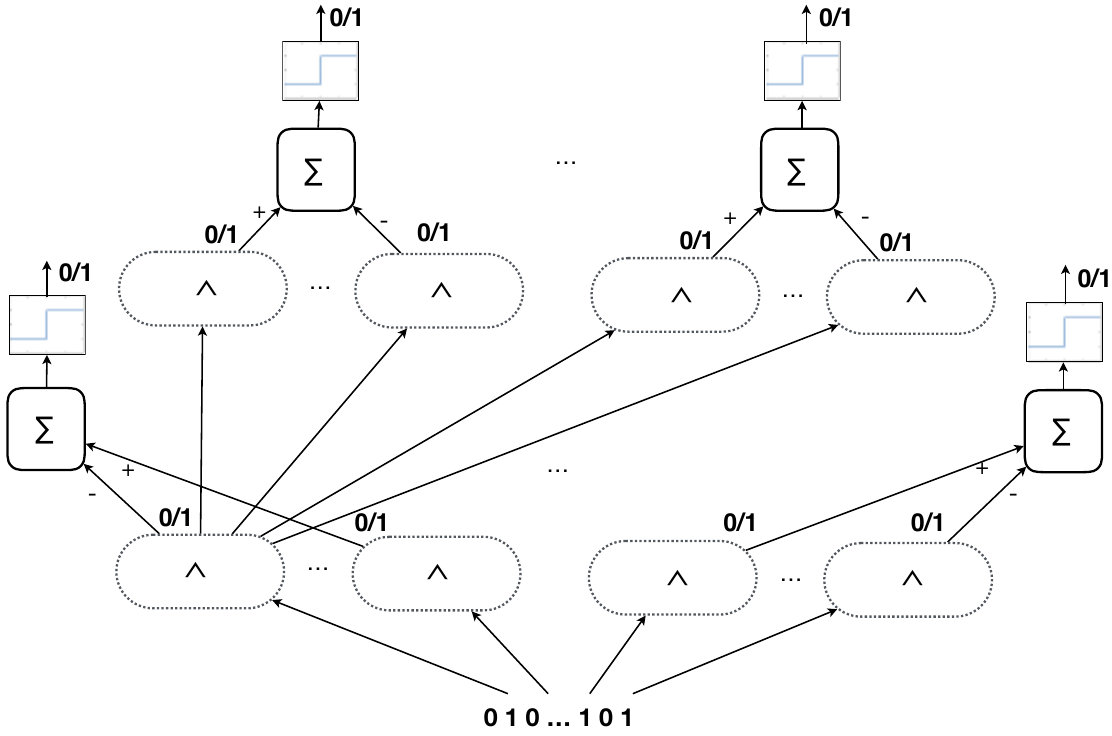}
\caption{The fully connected Deep Tsetlin Machine.}
\label{figure:deep_tsetlin_machine}
\end{figure}

Another architectural family is the Fully Connected Deep Tsetlin Machine \cite{Granmo2018b}, illustrated in Figure \ref{figure:deep_tsetlin_machine}. 
The purpose of this architecture is to build composite propositional formulas, combining the propositional formula composed at one layer into more complex formula at the next. As exemplified in the figure, we here connect multiple Tsetlin Machines in a sequence. The clause output from  each Tsetlin Machine in the sequence is provided as input to the next Tsetlin Machine in the sequence. In this manner we build a multi-layered system.  For instance, if layer $t$ produces two clauses $(P \land \lnot Q)$ and $(\lnot P \land Q)$, layer $t+1$ can manipulate these further, treating them as inputs. Layer $t+1$ could then form more complex formulas like $\lnot (P \land \lnot Q) \land (P \land \lnot Q)$, which can be rewritten as $(\lnot P \lor Q) \land (P \land \lnot Q)$.

One simple approach for training such an architecture is indicated in the figure. As illustrated, each layer is trained independently, directly from the output target $y$, exactly as described in Section \ref{sec:tsetlin_machine_game}. The training procedure is thus similar to the strategy Hinton et al. used to train their pioneering Deep Belief Networks, layer-by-layer, in 2006 \cite{Hinton2006}. Such an approach can be effective when each layer produces abstractions, in the form of clauses, that can be taken advantage of in the following layer.

\subsection{The Convolutional Tsetlin Machine}

\begin{figure}[!th]
\centering
\includegraphics[width=6.0in]{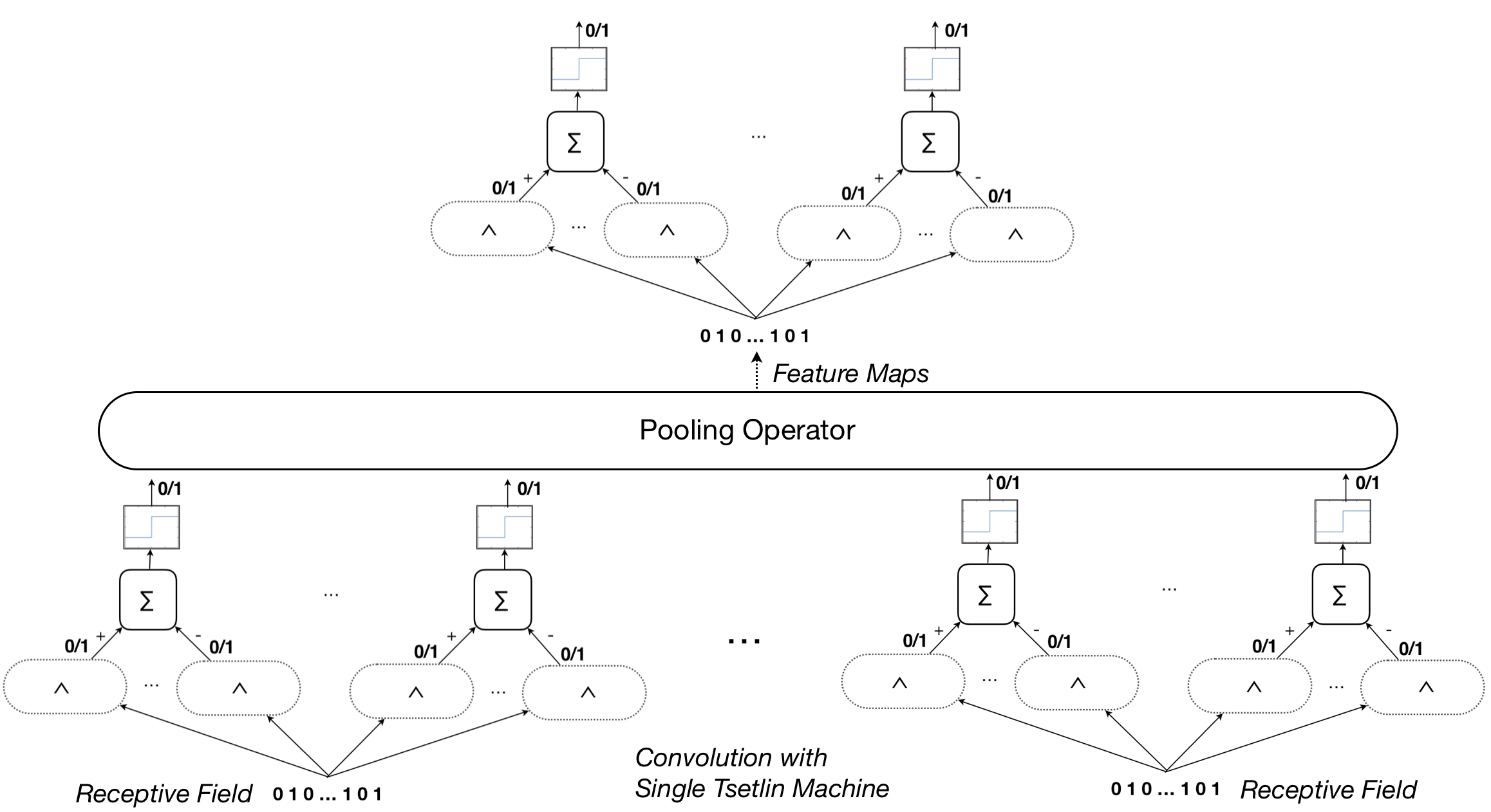}
\caption{The Convolutional Tsetlin Machine.}
\label{figure:convolutional_tsetlin_machine}
\end{figure}

We next demonstrate how self-contained and independent Tsetlin Machines can interact to build a Convolutional Tsetlin Machine \cite{Granmo2018d}, illustrated in Figure \ref{figure:convolutional_tsetlin_machine}.
The Convolutional Tsetlin Machine is a deep architecture based on mathematical convolution, akin to Convolutional Neural Networks \cite{LeCun1998}. For illustration purposes, consider 2D images of size $100 \times 100$ as input. At the core of a Convolutional Tsetlin Machine we find a kernel Tsetlin Machine with a small receptive field. Each layer $t$ of the Convolutional Tsetlin Machine operates as follows:
\begin{enumerate}
    \item A convolution is performed over the input from the previous Tsetlin Machine layer, producing one \emph{feature map} per output $y$. Here, the Tsetlin Machine acts as a kernel in the convolution. In this manner, we reduce complexity by reusing the same Tsetlin Machine across the whole image, focusing on a small image patch at a time.  
    \item The feature maps produced are then down-sampled using a pooling operator, in a similar fashion as done in a Convolutional Neural Network,  before the next layer and a new Tsetlin Machine takes over. Here, the purpose of the pooling operation is to gradually increase the abstraction level of the clauses, layer by layer.
\end{enumerate}
A simple approach for training a Convolutional Tsetlin Machine is indicated in the figure. In brief, the feedback to the Tsetlin Machine kernel is directly provided from the desired end output $y$, exactly as described in Section \ref{sec:tsetlin_machine}. The only difference is the fact that the input to layer $t+1$ comes from the down-scaled feature map produced by layer $t$. Again, this is useful when each layer produces abstractions, in the form of clauses, that can be taken advantage of at the next layer.

\subsection{The Recurrent Tsetlin Machine}

\begin{figure}[!th]
\centering
\includegraphics[width=4.0in]{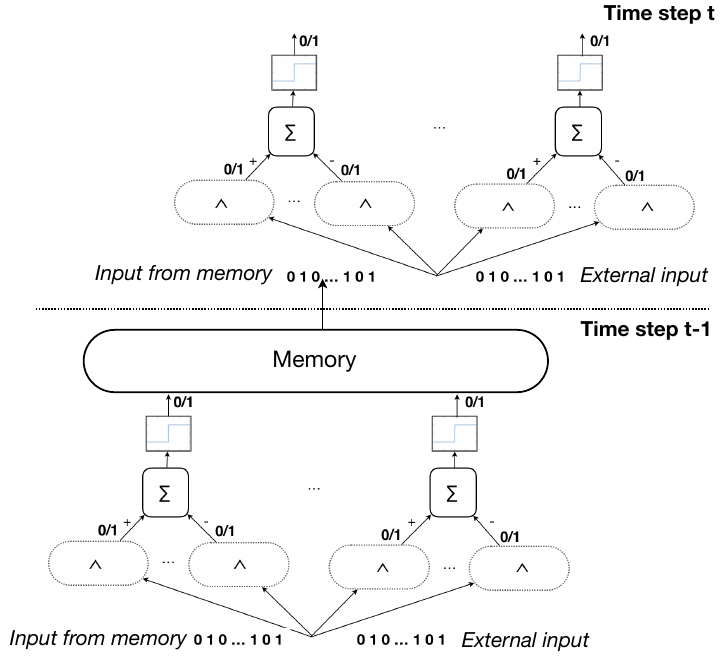}
\caption{The Recurrent Tsetlin Machine.}
\label{figure:recurrent_tsetlin_machine}
\end{figure}

The final example is the Recurrent Tsetlin Machine \cite{Granmo2018c} (Figure \ref{figure:recurrent_tsetlin_machine}). In all brevity, the same Tsetlin Machine is here reused from time step to time step. By taking the output from the Tsetlin Machine of the previous time step as input, together with an external input from the current time step, an infinitely deep sequence of Tsetlin Machines is formed. This is quite similar to the family of Recurrent Neural Networks \cite{Schmidhuber2015}.
Again, the architecture can be trained layer by layer, directly from the target output $y(t)$ of the current time step $t$. However, to learn more advanced sequential patterns, there is a need for rewarding and penalizing that propagate back in time. How to design such a propagation scheme is presently an open research question.

\section{Conclusion and Further Work}
\label{sec:conclusion}

In this paper we proposed the Tsetlin Machine, an alternative to neural networks. The Tsetlin Machine solves the vanishing signal-to-noise ratio of collectives of Tsetlin Automata. This allows it to coordinate millions of Tsetlin Automata. By equipping teams of Tsetlin Automata with the ability to express patterns in propositional logic, we have enabled them to recognize complex patterns. Furthermore, we proposed a novel decentralized feedback orchestration mechanism. The mechanism is based on resource allocation principles, with the intent of maximizing effectiveness of sparse pattern recognition capacity. This mechanism effectively provides the Tsetlin Machine with the ability to capture unlabelled sub-patterns.

Our theoretical analysis reveals that the Tsetlin Machine game have Nash equilibria that maps to propositional formulas that maximize pattern recognition accuracy. In other words, there are no local optima in the learning process, only global ones. This explains how the collectives of Tsetlin Automata are able to accurately converge towards complex propositional formulas that capture the essence of five diverse pattern recognition problems.

Overall, the Tsetlin Machine seems to be particularly suited for digital computers, being based on simple bit manipulation with AND-, OR-, and NOT gates. Both input, hidden patterns, and output can be expressed as bit patterns.

In our empirical evaluations on five benchmarks, the Tsetlin Machine provided competitive accuracy with respect to both Multilayer Perceptron Networks, Support Vector Machines, Decision Trees, Random Forests, the Naive Bayes Classifier and Logistic Regression. It further turns out that the Tsetlin Machine requires less data than neural networks, outperforming the Naive Bayes Classifier in data sparse environments. 

By demonstrating that the longstanding problem of vanishing signal-to-noise ratio can be solved, the Tsetlin Machine further provides a novel game theoretic framework for recasting the problem of pattern recognition. This framework can thus provide opportunities for introducing bandit algorithms into large-scale pattern recognition. It could for instance be interesting to investigate the effect of replacing the Tsetlin Automaton with alternative bandit algorithms, such as algorithms based on Thompson Sampling \cite{Thompson1933,Granmo2010f,May2012,Chapelle2011} or Upper Confidence Bounds~\cite{Auer2003}. 

The more advanced Fully Connected Deep Tsetlin Machine, the Convolution Tsetlin Machine, and the Recurrent Tsetlin Machine architectures also form a starting point for further exploration. These architectures can potentially improve pattern representation compactness and learning speed. However, it is currently unclear how these architectures can be most effectively trained.

Lastly, the high accuracy of the Tsetlin Machine, combined with its ability to produce self-contained easy-to-interpret propositional formulas for pattern recognition, makes it attractive for applied research, such as in the safety-critical medical domain.
 
\section*{Acknowledgements}
I thank my colleagues from the Centre for Artificial Intelligence Research (CAIR), Lei Jiao, Xuan Zhang, Geir Thore Berge, Darshana Abeyrathna, Saeed Rahimi Gorji, Sondre Glimsdal, Rupsa Saha, Bimal Bhattarai, Rohan K. Yadev, Bernt Viggo Matheussen, Morten Goodwin, Christian Omlin, Vladimir Zadorozhny (University of Pittsburgh), Jivitesh Sharma, and Ahmed Abouzeid, for their contributions to the development of the Tsetlin machine family of techniques. I would also like to thank our House of CAIR partners, Alex Yakovlev, Rishad Shafik, Adrian Wheeldon, Jie Lei, Tousif Rahman (Newcastle University), Jonny Edwards (Temporal Computing), Marco Wiering (University of Groningen), Christian D. Blakely (PwC Switzerland), Adrian Phoulady, Anders Refsdal Olsen, Halvor Smørvik, and Erik Mathisen for their many contributions.

\section*{Code Availability}
Source code and datasets for the Tsetlin Machine, available under the MIT Licence, can be found at \href{https://github.com/cair/TsetlinMachine}{https://github.com/cair/TsetlinMachine} and \href{https://github.com/cair/pyTsetlinMachine}{https://github.com/cair/pyTsetlinMachine}.

\section*{Data Availability}
The datasets generated during and/or analysed during the current study are available from the corresponding author on reasonable request. 

\bibliographystyle{IEEEtran}
\bibliography{Granmo}

\end{document}